\documentclass[sigconf]{acmart}

\copyrightyear{2022}
\acmYear{2022}
\setcopyright{acmlicensed}\acmConference[SIGIR '22]{Proceedings of the 45th
International ACM SIGIR Conference on Research and Development in
Information Retrieval}{July 11--15, 2022}{Madrid, Spain}
\acmBooktitle{Proceedings of the 45th International ACM SIGIR Conference on
Research and Development in Information Retrieval (SIGIR '22), July 11--15,
2022, Madrid, Spain}
\acmPrice{15.00}
\acmDOI{10.1145/3477495.3531957}
\acmISBN{978-1-4503-8732-3/22/07}

\AtBeginDocument{%
  \providecommand\BibTeX{{%
    \normalfont B\kern-0.5em{\scshape i\kern-0.25em b}\kern-0.8em\TeX}}}

\usepackage{balance}
\usepackage{graphicx}
\usepackage{amsmath}
\usepackage{multirow}
\usepackage{threeparttable}
\usepackage{booktabs}
\usepackage{hyperref}
\usepackage{dsfont}
\usepackage{caption} 
\usepackage{bbold}
\usepackage{enumitem}
\definecolor{purple}{RGB}{112,48,160}
\definecolor{ocean}{RGB}{2,154,152}
\definecolor{blue}{RGB}{31,119,180}
\usepackage{comment}
\newcommand\modelname{\textbb{COSPLAY}}

\begin{document}
\fancyhead{}

\title{\modelname{}: Concept Set Guided Personalized Dialogue Generation Across Both Party Personas}

\author{Chen Xu}
\affiliation{%
  \institution{Beijing University of Technology}
  \country{}
}
\email{chenxu05037@gmail.com}

\author{Piji Li}
\affiliation{%
 \institution{Tencent AI Lab}
 \country{}}
\email{lipiji.pz@gmail.com}

\author{Wei Wang}
\affiliation{%
 \institution{Tsinghua University}
 \country{}}
 \email{w-w16@mails.tsinghua.edu.cn}

\author{Haoran Yang}
\affiliation{%
 \institution{The Chinese University of Hong Kong}
 \country{}}
\email{hryang@se.cuhk.edu.hk}

\author{Siyun Wang}
\affiliation{%
 \institution{University of Southern California}
 \country{}}
 \email{siyunwang1027@gmail.com}

\author{Chuangbai Xiao}
\affiliation{%
  \institution{Beijing University of Technology}
 \country{}}
 \email{cbxiao@bjut.edu.cn}


\begin{abstract}
Maintaining a consistent persona is essential for building a human-like conversational model. However, the lack of attention to the partner makes the model more egocentric: they tend to show their persona by all means such as twisting the topic stiffly, pulling the conversation to their own interests regardless, and rambling their persona with little curiosity to the partner.
In this work, we propose \modelname{} 
(\textbf{CO}ncept \textbf{S}et guided \textbf{P}ersona\textbf{L}ized dialogue generation \textbf{A}cross both part\textbf{Y} personas)
that considers both parties as a ``team'': 
expressing self-persona while keeping curiosity toward the partner, leading responses around mutual personas, and finding the common ground. 
Specifically, we first represent self-persona, partner persona and mutual dialogue
all in the concept sets. Then, we propose the \textbf{Concept Set} framework with a suite of knowledge-enhanced operations to process them such as set algebras, set expansion, and set distance. 
Based on these operations as medium, we train the model by utilizing 1) concepts of both party personas, 2) concept relationship between them, and 3) their relationship to the future dialogue.
Extensive experiments on a large public dataset, Persona-Chat, demonstrate that our model outperforms state-of-the-art baselines for generating less egocentric, more human-like, and higher quality responses in both automatic and human evaluations. 
\end{abstract}



\maketitle

\section{Introduction}
\label{sec:intro}

\begin{figure}[t]
\centering
\includegraphics[width=\columnwidth]{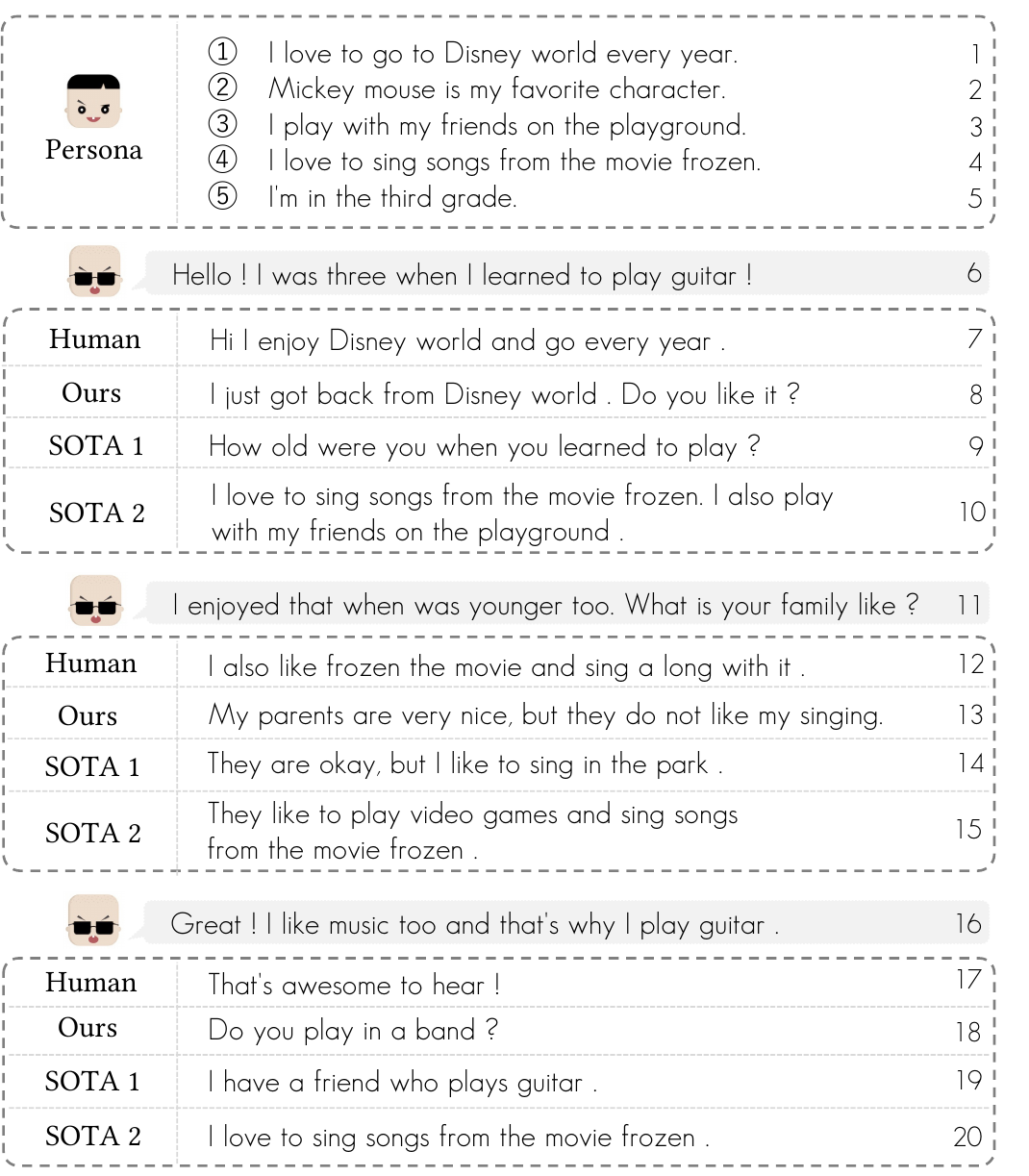}
\caption{Case study for personalized dialogue generation.}
\vspace{-4mm}
\label{fig:case}
\end{figure}

Building a more human-like dialogue system has been an important topic in artificial intelligence, where one of the main challenges is to maintain a consistent persona such as age, gender, occupations, etc.~\cite{li2016persona, shum2018eliza, wang2018chat, persona-huang, zhou2018commonsense, qian2018assigning, kottur2017exploring}. 
More recently, a more direct approach, defining the persona as several profile sentences, was proposed with a novel dataset \textsc{Persona-Chat}~\cite{zhang2018personalizing}. This task required models to generate responses consistent with several persona sentences (Figure~\ref{fig:case}) ~\cite{zhang2018personalizing}. Due to the flexibility for users to describe complicated, compositional profile and to choose whoever they want to talk with, the dataset sparked a wide range of interest in developing methods on generating the persona-consistent responses
~\cite{zhang2018personalizing, dinan2019the, Song_RCDG_2020, yavuz2019deepcopy, wolf2019transfertransfo, golovanov2019large, mazar2018training, zheng2020pre}.
To better control the consistency, some additional work were made to further enhancing the persona understanding of the model~\cite{liu2020you, song-etal-2021-bob, song2020generate}: 
~\citet{song-etal-2021-bob} introduced unlikelihood training with non-dialogue inference data to strengthen consistency understanding. 
~\citet{liu2020you} utilized a fine-grained consistency reward model trained by negative sampling to reinforce consistency understanding of the model.
~\citet{song2020generate} integrated a matching model to detect and delete inconsistent words and further rewrite it to a persona-consistent one.

However, as an increasingly amount of efforts have been devoted to encourage persona-consistent responses, the agents tend to be more and more \textbf{egocentric}: they tend to demonstrate their own persona information by all means while show less interests about the partner's. 
Here we take the examples in Figure~\ref{fig:case} to further illustrate this phenomenon. In line 11, in response to the question ``what is your family like?'', one baseline model replies ``They are okey, \textbf{but} I like to sing in the park''. We can see that in order to express its own persona ``I love to sing songs'' eagerly, the model hastily and bluntly twists the topic by using an inappropriate transitional word ``but'', resulting in an \textbf{illogical} response. 
Considering another answer ``\textbf{They} like sing songs'' where the baseline who even cannot be bothered to transit the topic just graft its own persona ``I love to sing songs'' to their family, which indicates that over focusing its own persona instead makes the response \textbf{inconsistent}.
There is another example in line 16. When a user says ``Great! I like music too and that's why I play guitar'', we, as human beings, can easily feel that he is very looking forward to further interacting and resonating with us in terms of his persona ``music'' and ``guitar''. The expected responses should be ``That's awesome to hear!'' or ``Do you play in a band?''. However, the baselines ignore the user's emotion and just pull the conversation to their own personas again by starting the response with first-person pronouns: ``\textbf{I love} to sing songs'' or ``\textbf{I have} a friend'' without showing any interests or further questions about the partner persona because giving the partner more opportunities to express ``sacrifices'' the chances to show its own persona. However, this also ``sacrifices'' the \textbf{user experience} and the \textbf{model interactivity}.

We argue the reason for egocentrism is that \textbf{without giving enough attention to the partner, expression of self-persona may sacrifice that of partner’s}. 
In everyday life conversations, referring the examples in Figure~\ref{fig:case} again, we do a great job in balancing self expression, asking questions to the partner (e.g. ``I just got back from Disney world. \textbf{Do you} like it?'' in line 8) and giving the partner more opportunities to express persona (e.g. ``\textbf{Do you} play in a band?'' to respond  ``I play guitar'' in line 18). In addition, with the partner's persona acquired, we try to lead the conversation around both self and partner's persona by bridging them (e.g. ``My \textbf{parents}[$\leftarrow$ family] are very nice, but they do not like my \textbf{singing}[$\leftarrow$ sing songs].'' in line 13) or finding the common ground between them (e.g., ``music''/``guitar'' $\rightarrow$ 
\textbf{band} $\leftarrow$ ``sing''/``songs'' in line 18), which shows that what we know from the partner's persona decides what information to choose from our persona when building the response. From all examples above, we can see that giving enough attention to the partner's persona plays an important role in generating less egocentric responses and making a big contribution to generate realistic conversations.







In this paper, we view self and partner as a ``team'' to capture this global consistency with the proposed model \modelname{} (\textbf{CO}ncept \textbf{S}et guided \textbf{P}ersona\textbf{L}ized dialogue generation \textbf{A}cross both part\textbf{Y} personas) that can 1) balance ``answering'' and ``\textbf{asking}'', expressing itself while being curious about the partner's persona which gives the partner more opportunities to express, 2) balance ``speaking'' and ``\textbf{listening}'', leading responses around mutual personas and finding the common ground.

Motivated by this, our research starts by asking: \textbf{How to encourage the model to ask question frequently and give the partner more chance to express persona in a generative way?} Naively inserting general predefined question templates makes response stiff and unable to ask concrete questions in different context. What we only know is that if everyone is willing to give the partner more chances to express persona, then the persona information of both parties will be recalled in the future dialogue. Motivated by this, we train the model in a lookahead way. Specifically, we let another \modelname{} to ``cosplay'' the partner with different persona to complete a dialogue with the trainable one and we attempt to maximize the intersection set between the concepts of this future dialogue and concepts of both personas. However, after we do that, the models run in the opposite direction: they no longer care about what the partner says and only rambles their persona information. Obviously, the model finds a bad way to get a big reward. In order to help the model in distinguishing this bad way from our expected one, we join the coherence score. At this time, with the combined effects, the model runs towards the expected way because it finds that adding questions towards the partner properly not only as a trigger to let the partner recall more but also as a bridge to keep the coherency. Hence, we propose the \textbf{Mutual Benefit Reward}, a combined score of recall and coherent scores.

After calculating the recall score, another problem appears: due to the persona sparsity, some personas highly correlated concepts in the future dialogue do not count in the intersection, thus not contributing to the recall score. To address this, we propose \textbf{Concept Set} framework with a suit of knowledge-enhanced operations (set algebras, set expansion, set similarity) dedicated to operate concepts, which assigns relevancy (distance) to the correlated concepts.

Next, the second question is: \textbf{How to fuse both party personas and automatically find the common ground?} Supported by Concept Set, \modelname{} uses a concept copy mechanism trained to copy information from both sides (concepts of self-persona and the partner's one shown in the context). Furthermore, to also explore and introduce concepts connecting them, \textbf{Common Ground Reward} is designed where we regard both party personas as two constraints and simultaneously minimize the concept set distances from the future dialogue to each of them in a geometrical way.


In summary, our key contributions are as follows:
\begin{itemize}[leftmargin=3.4mm]

\item We take the first step towards addressing the problem of egocentrism by applying ConceptNet in the personalized dialogue generation and propose \modelname{} to achieve a new consistency paradigm: being curious about the partner, leading conversation around mutual personas, and finding the common ground. 
\item We explore a new way for the usage of ConceptNet and propose the \textbf{Concept Set} framework: modeling personas and dialogues all in concept sets and using set operations to calculate their relationships.
Driven by Concept Set, \modelname{} utilizes both concept set and concept set relationships to guide the model 
and operates knowledge in parallel computing, without the need to construct a sub-graph, search in the whole graph, or do one or multi-hop reasoning during the generation. 
\item We find an efficient way to encourage model to be curious about the partner by \textbf{Mutual Benefit Reward} that utilizes the concept relationship between the future generated dialogue and both personas.
Moreover, to encourage the model to find the common ground, 
we design \textbf{Common Ground Reward} by viewing both personas as two constraints and make use of the concept relationship between them to train the model.


\item Experiments on \textsc{Persona-Chat} demonstrate that our model \modelname{}, by giving more attention to the partner, can generate less egocentric, more human-like, and higher quality responses.
\end{itemize}

\section{Related Work}

\subsection{Personalized Dialogue System}
Open-domain dialogue systems has long been an important and challenging task in the field of AI. Although neural response generation models have achieved promising results~\citep{shang2015neural,zhao2017learning,zhang2019dialogpt}, they are still unsatisfactory. One possible reason is that they lack a consistent personality. 
\citet{li2016persona} is the first to propose a personalized dialogue system for improving response consistency.
\citet{zhang2018personalizing} introduced the \textsc{Persona-Chat}. The dataset facilitated the training of chatbots with predefined textually described persona sentences and sparked a wide range of interest in developing both retrieval and generative based personalized dialogue models.

In the line of retrieval-based methods~\cite{zhang2018personalizing, gu2019dually, gu2021partner}, a relevant work was done by \citet{gu2021partner}. They provide an empirical study on retrieval-based methods and find that it is helpful to utilize both persona sentences into response selection. However, the main drawback of this line of approaches is that selecting responses from a pre-constructed pool cannot satisfy different persona settings such as diverse combinations of occupation and location. Therefore, to produce an informative while consistent responses, a better way is to build generative dialogue models.
In the line of generative-based methods~\cite{zhang2018personalizing, Song_RCDG_2020, yavuz2019deepcopy, wolf2019transfertransfo, golovanov2019large, xu2020neural, dinan2019the, liu2020you, bi2022amodel, tian2021learning}, with the advancement of the pre-training~\cite{radford2018improving, radford2019language, vaswani2017attention}, two-stage methods are applied for personalized dialogue generation such as fine-tuning a pre-trained model on the \textsc{Persona-Chat}~\cite{wolf2019transfertransfo, golovanov2019large, dinan2019the, liu2020you}.
Our work also follows this line where a relevant work was done by~\citet{liu2020you}. They train a receiver to reinforce the mutual persona understanding. Specifically, the receiver is trained by negative samples where self utterances are paired with partner persona to obtain fine-grained consistency understanding ability.
We borrow the idea that incorporating the partner persona by self-play with several differences in motivation and methodology: 
1) The partner persona sentences in their work are used as negative samples to enhance model consistency understanding while we incorporate both party personas as positive samples to alleviate model egocentrism;
2) Considering personas of both sides simultaneously (not separately) and equally (not as negative samples) enables us to further utilize the relationship between them;
3) We incorporate commonsense knowledge to calculate reward without training implicitly reward models.

\subsection{Knowledge Enhanced Text Generation}

Incorporating external knowledge is essential for text generation to augment the limited information in the context or better achieve tasks~\cite{zhou2018commonsense, speer2017conceptnet, guan2019story, ji2020language, zhou2021commonsense, xu2021change, li2020knowledge}.
As one of the most common knowledge graph, ConceptNet~\cite{speer2017conceptnet} is a large-scale multilingual semantic graph that describes general human knowledge in natural language. Each node can be a word or a phrase, and the edges represent the semantic relations between nodes with the confidence score as weights. To enhance emotion perception and expression, ~\citet{li2020knowledge} inject concepts with higher emotion intensity values from ConceptNet into the model in empathetic dialogue generation.
\citet{zhou2018commonsense} employ graph attention embedding to encode sub-graphs which contains neighbor concepts on ConceptNet in dialogue generation.
\citet{ji2020language} enables pre-trained models with dynamic multi-hop reasoning on the sub-graph.
\citet{guan2019story} incorporated one-hop knowledge graph for concepts in the story context for the incremental encoding with multi-source attention.

Compared to ours, we follow common practice of utilizing ConceptNet to expand the concepts in the context. 
However, some key innovations that set \modelname{} aside from theirs are as follows: 1) we take the first step towards applying ConceptNet for the problem of egocentrism in the personalized dialogue generation by linking both parties, 2) we also use ConceptNet to calculate the relationships (persona-persona and persona-dialogue) as rewards to train the model, and 3) we explore a new way for the usage of ConceptNet by combining set theory, linear algebra, and knowledge graph. All operations over concepts have been framed as matrix calculations, joining parallel computing along with the neural networks.


\section{Methodology}
\begin{figure*}[!t]
\vspace{-3mm}
\centering
\includegraphics[width=\textwidth]{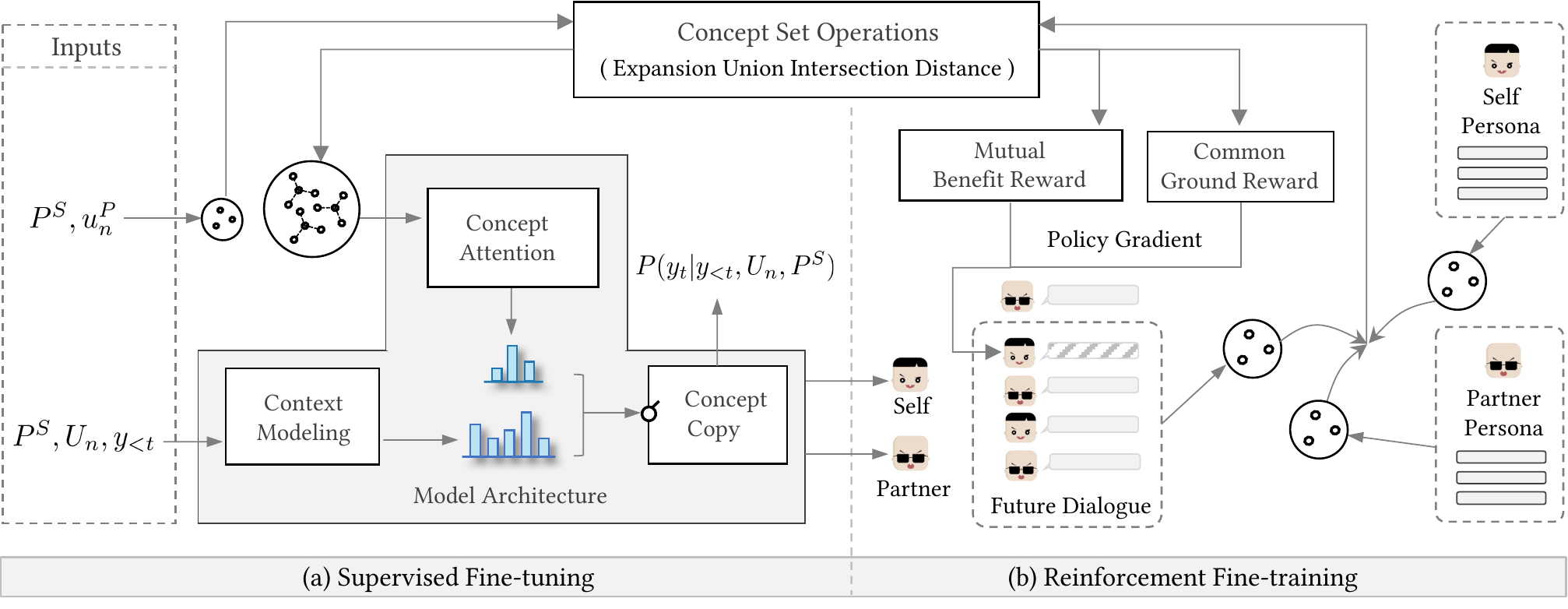}
\vspace{-6mm}
\caption{The overview of method. Our model is guided by the concept set in both supervised and reinforcement phases. In the supervised phase (a), it's guided by the concept set itself: \modelname{} learns to copy concepts from both of self-persona $P^S$ and partner persona shown in the context $u_n^P$.
After supervised phase, we pair another \modelname{} as the partner with a different persona to complete a future dialogue together. Next, in the reinforcement scenario (b), our model is guided by the concept set relationships: \modelname{} learns to maximize two global rewards across both parties. The rewards are applied to the current response (striped), based on the relationships between self-persona, partner persona, and future dialogue.}
\label{fig:overview}
\end{figure*}

\subsection{Overview}
\label{sec:overview}

\label{sec:PF}
A $N$-turn dialogue is defined as a sequence of utterances $U=\{u^P_1, u^S_1,...,u^P_N, u^S_N\}$, where $u^P_n$ denotes the utterance that the \textbf{P}artner (user) says and $u^S_n$ denotes the utterance that the model (\textbf{S}elf) responds to the partner at $n$-th turn. When they first meet each other, the dialogue started by the partner and the model is expected to generate utterance according to its $L$ self-persona sentences $P^S=\{p_1^S,...,p_L^S\}$. 
Therefore, the personalized dialogue model is trained to model the response $y={u}^S_n$ conditioned on previous utterances $U_n=\{u^P_1, u^S_1,...,u^P_{n}\}$ and its self-persona sentences $P^S$:
\begin{equation}
    u^S_n = \texttt{argmax}_yP(y|U_n, P^S)
\end{equation}
Figure~\ref{fig:overview} gives an overview of our method. Our model is trained by two stages: supervised fine-tuning and reinforcement fine-training. 
In supervised fine-tuning phase (Figure~\ref{fig:overview}a), three modules work together to fuse concepts across both personas into the response:
\begin{itemize}[leftmargin=3mm]
\item \textbf{Context Modeling} with pre-trained transformer encodes self-persona text, dialogue history, and generated words so far.
\item \textbf{Concept Attention} selects the best concepts in the expanded concept set extracted from both party personas at each time step.
\item \textbf{Concept Copy} is responsible for the gate control, letting concepts flow in at the appropriate time during the generation.
\end{itemize}
In reinforcement fine-training phase (Figure~\ref{fig:overview}b), 
we design two rewards based on the concept relationships to guide the model:
\begin{itemize}[leftmargin=3mm]
\item \textbf{Mutual Benefit Reward} encourages the model to be more curious about the partner and to both ask and answer questions by utilizing the relationship between future dialogue and both personas. It is also helpful to encourage the model to copy both party persona concepts because of its component recall score. 
\item \textbf{Common Ground Reward} encourages the model to find the common ground by utilizing the relationship between them.
\end{itemize}
Across both phases, all the operations over concepts are supported by a suite of knowledge-enhanced \textbf{Concept Set Operations}.



\subsection{Concept Set Construction and Operations}
\label{sec:conceptSet}

\subsubsection{Concept Set Construction}
\label{sec:cs}
We define a \textbf{c}oncept set (Figure~\ref{fig:operations}a) as a logical vector $\mathbf{c}^A\in \{0, 1\}^{|V|}$, which can be viewed as a sparse mask vector over a concept vocabulary $V$. $\mathbf{c}^A$ is the vector version of the set $A = \{{v_i}\in V \ |\ \mathbf{c}_i=1 \}$ where $i$ is the index of $V$. The number of concepts in set can be calculated by its L1 norm $ \lVert \mathbf{c}^A\rVert_1$. 

A concept set is extracted from a piece of text in the process of filtering stop words, lematizing/stemming, and maintaining the words only in the concept vocabulary $V$ as the set elements. The concept vocabulary is borrowed from~\citet{zhong2021keyword}. In the study of keyword-guided conversation, they collected a keyword vocabulary whose elements are selected from single content words in the ConceptNet. We modified this vocabulary to drop some common words (e.g., like) and words not in our language model (LM) vocabulary, and transform words to their basic form (e.g, dogs$\rightarrow$dog) in order to keep the dimension of the concept set and the concept distance matrix (\hyperref[sec:matrix]{\textsection\ref{sec:matrix}}) scalable for parallel computing.





\subsubsection{Concept Distance Matrix}
\label{sec:matrix}
\begin{figure*}[t]
\centering
\vspace{-0.83mm}
\includegraphics[width=.99\textwidth]{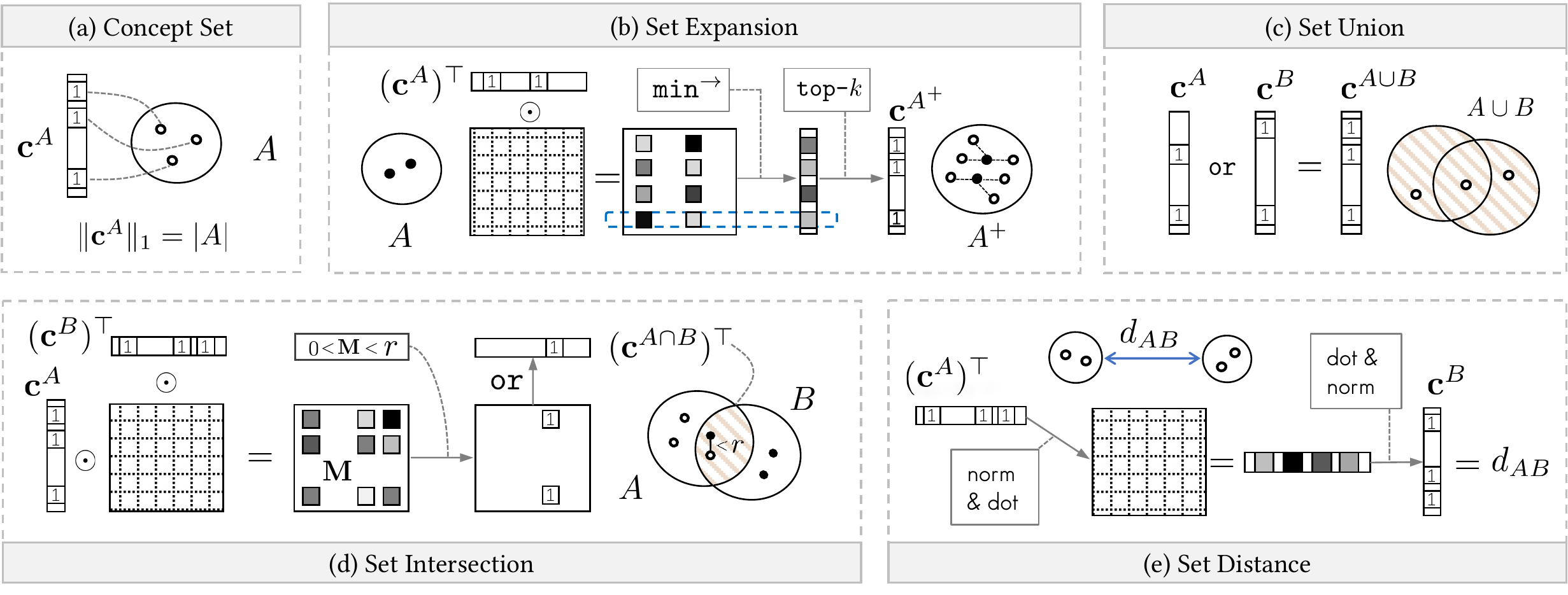} 
\vspace{-2mm}
\caption{Concept set operations. The $k$ and $r$ are parameters to control the concept generalization strength.}
\label{fig:operations}
\end{figure*}

The concept distance matrix (grid square in Figure~\ref{fig:operations}) is calculated from the ConceptNet~\cite{speer2017conceptnet}. We define the concept distance matrix as a symmetric matrix $\mathbf{D}=[d_{ij}]_{|V|\times |V|}$
where each element $d_{ij}=d_{ji}\in [0.001\texttt{,} \texttt{MAX}]$ is the weighted length of the shortest path between the concept $v_i$ and the concept $v_j$.
Note that the minimum value $d_{ii}$ is set as 0.001 because we distinguish it from zero (this is important when we do mask operation $\odot$ over $\mathbf{D}$).

\subsubsection{Set Expansion Operation}
Set expansion operation (Figure~\ref{fig:operations}b) uses the concepts in the original set $A$ (solid dots) to query their related ones (hollow dots). The process is formulated as follows:
\begin{equation}\label{eq:expansion}
    \texttt{Expa}(A;k) = \texttt{topk}(\texttt{min}({(\mathbf{c}^A)}^{\top} \odot \mathbf D), k)
\end{equation}
where $\odot$ (mask operation) denotes the element-wise multiplication with broadcasting and 
$\texttt{min}$ denotes the min pooling operation on row dimension, which can be viewed as classifying each concept $v_i$ to its nearest cluster in the set $A$ by retaining the minimum distance.
Finally, we obtain $k$ (parameter) concepts ordered by shortest distances in the final concept set.

\subsubsection{Set Union Operation}
The union of two concept sets (Figure~\ref{fig:operations}c) contains all the elements stored in either set, which is same as the one in traditional set theory: 
\begin{equation}\label{eq:union}
        \texttt{Union}(A,B)\\
          = \texttt{or}(\mathbf{c}^A, \mathbf{c}^B)
\end{equation}
where $\texttt{or}$ denotes the logical or operation over two logical vectors.
\subsubsection{Set Intersection Operation}
We define the intersection set of two concept sets as $A\cap B={\{v_i\in B\ |\ \exists \ v_j\in A\ni {d}_{ij} < r\}}$. Note that this intersection set is not the traditional one in set theory because of the existence of parameter $r$ that enables non-exact matching (if $r\rightarrow0$, then the operation will degenerate into the traditional intersection operation). Therefore, it is not commutative because we define this intersection as the subset of $B$. The intersection set $A\cap B$ is obtained by the concept set intersection operation (Figure~\ref{fig:operations}d):
\begin{equation}\label{eq:intersection}
    \begin{aligned}
        \texttt{Inter}( A,B;r) &= \texttt{or}(0<\mathbf M < r)\\
        \mathbf M &= \mathbf c^A \odot \mathbf D \odot (\mathbf c^B)^{\top} 
    \end{aligned}
\end{equation}
where $\mathbf M$ is the masked $\mathbf D$, $r$ is a parameter as the threshold indicating the degree of similarity, and $\texttt{or}$ represents logical or operation on the column dimension of the logical matrix $(0<\mathbf M < r)$.
\subsubsection{Set Distance Operation}
The distance operation (Figure~\ref{fig:operations}e) is to measure the distance between any two concept sets:
\begin{equation}\label{eq:distance}
        \texttt{Dist}(A,B)\\
          = \frac{{(\mathbf c^A)}^{\top}}{\lVert \mathbf c^A\rVert_1} \cdot \mathbf{D} \cdot \frac{\mathbf c^B}{\lVert \mathbf c^B\rVert_1}
\end{equation}
where the intermediate result of $(\frac{{(\mathbf c^A)}^{\top}}{\lVert \mathbf c^A\rVert_1} \cdot \mathbf{D})$ is the distance vector representing the distances from $A$ to each concepts. Then the dot product between this distance vector and the $\frac{\mathbf c^B}{\lVert \mathbf c^B\rVert_1}$ means that only the distances to the concepts in $B$ are considered.

\subsection{Concept Set Guided Response Generation}
\label{sec:sl}
\subsubsection{Context Modeling}
\label{sec:cm}
We adopt the GPT-2 model~\cite{radford2019language}, a $L$ transformer blocks stacked decoder with a classifier at the sequence tail.
The input to the model is the concatenation of self-persona sentences $P^{S}$, history utterances $U_n$, target response $y=u^S_{n}$ and some special tokens: $(P^S, [\texttt{SEP}], U_n, \texttt{[SEP]}, y, \texttt{[CLS]})$.
At each time step, we compute the probability of the next token as follows:
\begin{equation}
    h_t^0 = e_t + p_t,
\end{equation}
\begin{equation}\label{eq:ds}
    h_t^l = \texttt{Block}(\textbf{H}_{\leq t}^{l-1})
\end{equation}
\begin{equation}\label{eq:lm}
    P_t^{\texttt{LM}} = \texttt{softmax}(\text{W}_{\texttt{LM}}\cdot h_t^{L}+b)
\end{equation}
where $e_t$ and $p_t$ are the token embedding and the position embedding respectively, $\texttt{Block}$ is the transformer block with masked self-attention, and $h_t^{L}$ is the final hidden state at the $t$-th time step.

\subsubsection{Concept Attention}
\label{sec:ca}
This module assigns attention probabilities over the expanded concept set.
The expanded set is initially extracted from the persona text across both parties. That is, both of the self-persona sentences $P^{S}$ and the partner last utterance $u_n^P$ reflecting its persona information are collected (union operation) and expanded (expansion operation) by the concept set operations:
\begin{equation}
    \mathbf{c}^{a} = \texttt{Expa}(\texttt{Union}(\mathbf{c}^{P^S}, \mathbf{c}^{u_n^P});\ k) \\
\end{equation}
where $\mathbf{c}^{P^S}$ and $\mathbf{c}^{u_n^P}$ are two concept sets extracted from ${P^S}$ and $u_n^P$ respectively, $\texttt{Expa}$ (Eq.~\ref{eq:expansion}) and $\texttt{Union}$ (Eq.~\ref{eq:union}) are two Concept Set operations, and the expanded set $\mathbf{c}^a$ with $k$ concepts serves as a utterance-level guide waiting to be copied. Note that though the concepts of both parties are 
fused into one set, our model will not confuse them since the self-persona sentences $P^{S}$ are still fed into the Context Modeling module (\hyperref[sec:cm]{\textsection\ref{sec:cm}}).
Next, in order to keep the generated text grammatically sound, the re-stemming/lemmatization operation first maps each element in set $\mathbf{c}^a$ with basic form (explained in \hyperref[sec:cs]{\textsection\ref{sec:cs}}) to its correct form with the largest $P_t^{\texttt{LM}}$ (Eq.~\ref{eq:lm}) in its word group (sing, singing) before being copied: $\mathbf{c}^{A}_t = \texttt{re(} c^a\texttt{,} P_t^{\texttt{LM}}\texttt{)}$.  $\mathbf{c}^{A}_t$ can be viewed as a stretched version $\mathbf{c}^a$ over the GPT-2 vocabulary but with the same number of elements ($\lVert \mathbf{c}^A_t\rVert_1=\lVert \mathbf{c}^a\rVert_1$). 
Finally, the attention distribution~\cite{bahdanau2015neural} over the stretched concept set $\mathbf{c}^{A}_t$ at each time step $t$ is calculated as follows:
\begin{equation}
    \mathbf{E}^{\mathbf{c}}_t = \mathbf{c}^A_t\odot \mathbf{E}
\end{equation}
\begin{equation}
    P_t^{\mathbf{c}} = \texttt{softmax}_{\mathbf{c}^{A}=1}(\mathbf{E}^\mathbf{c}_t \cdot h_t^{L})
\end{equation}
where $\mathbf{E}$ is the word embedding matrix and $h_t^L$ represents the decoder states (Eq.~\ref{eq:lm}). 
Next, the distribution is used to produce the concept set states $h^\mathbf{c}_t$, the weighted sum of its concept embeddings:
\begin{equation}
    h^\mathbf{c}_t = \sum_{i} (P_t^{\mathbf{c}})_{i} \times e^\mathbf{c}_i
\end{equation}

\begin{figure}[!t]
\vspace{+0.33mm}
\centering
\includegraphics[width=0.9\columnwidth]{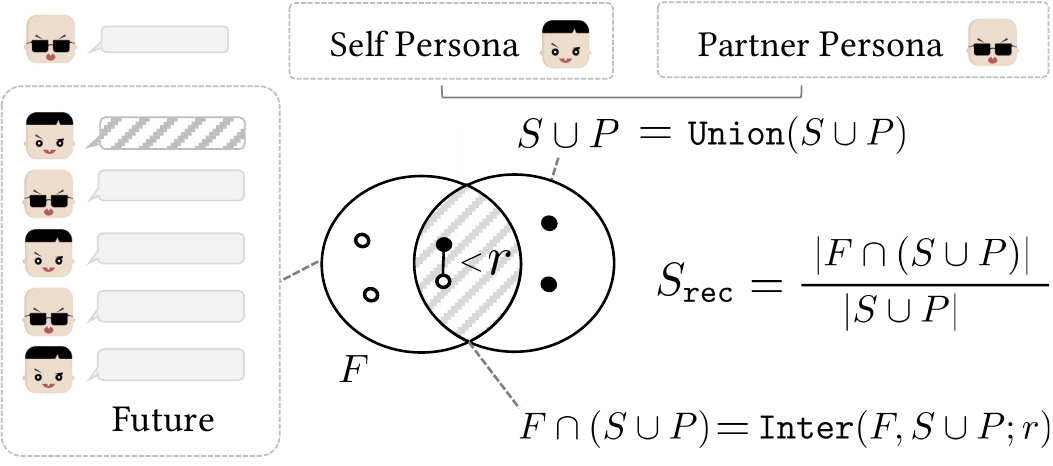}
\vspace{-1.72mm}
\caption{Persona recalls score of future dialogue.
The striped response is scored as an example.
The score can be viewed as a fraction of concepts in the union of self and partner persona concepts $S\cup P$ (closed dots) which are covered by the concepts in the future dialogue $F$ (open dots).}
\label{fig:recall}
\end{figure}


\subsubsection{Concept Copy}
\label{sec:ccm}
This module oversees the gate, letting concepts flow in at the appropriate time. Specifically, we use a soft gate probability $p^{\texttt{gate}}_t$ which denotes whether to copy a concept from the set. The gate controls the weight of the two distributions, which is similar to the copy mechanism~\cite{gu2016incorporating, see2017get, ji2020language}:
\begin{equation}
    p^{\texttt{gate}}_t = \sigma (\text{W}_1\cdot h^\mathbf{c}_t + \text{W}_2\cdot h_t^{L} + b)
\end{equation}
Finally the output distribution is the linear combination of LM distribution and the distribution over the concept set:
\begin{equation}
    \begin{aligned}
    P_{t} = p^{\texttt{gate}}_t \cdot P_t^{\mathbf{c}} + (1-p^{\texttt{gate}}_t) \cdot P_t^{\texttt{LM}}
    \end{aligned}
\end{equation}
and the generation loss for the response is calculated as follows: 
\begin{equation}
    \mathcal L_{\texttt{gen}} = -\sum_{t}\texttt{log}P_t
\end{equation}
Besides the generation loss, we further add two auxiliary losses: $\mathcal L_{\texttt{gate}}$ to supervise
the probability of selecting an element from the concept set or a generic word and $\mathcal L_{\texttt{next}}$ provided by the classifier built on top of $\texttt{[CLS]}$ trained by negative sampling to discriminate whether the response is the next utterance of the given context.
Both auxiliary loss functions take the form of cross-entropy and the final loss is the combination of these three losses:
\begin{equation}
    \mathcal L = \mathcal L_{\texttt{gen}} + \alpha_1\mathcal L_{\texttt{gate}} + \alpha_2\mathcal L_{\texttt{cls}}
\end{equation}

\subsection{Concept Set Guided Reinforcement Fine-Training}
\label{sec:FCG-RL}

In supervised fine-tuning, copying concepts from both persona information (the response persona sentences and the partner persona shown in context) can be viewed as an entrance to fuse both-party concepts into the response, which is used to generate responses mentioning mutual persona information simultaneously. 
In this section, we further introduce two rewards build on the concept set relationships between self-persona, partner persona, and future generated dialogue. These rewards are used to fine-training, which completely improve the model's attention to the partner while generating responses via reinforcement learning~\cite{williams1992simple}.

\begin{figure}[!t]
\centering
\includegraphics[width=.9\columnwidth]{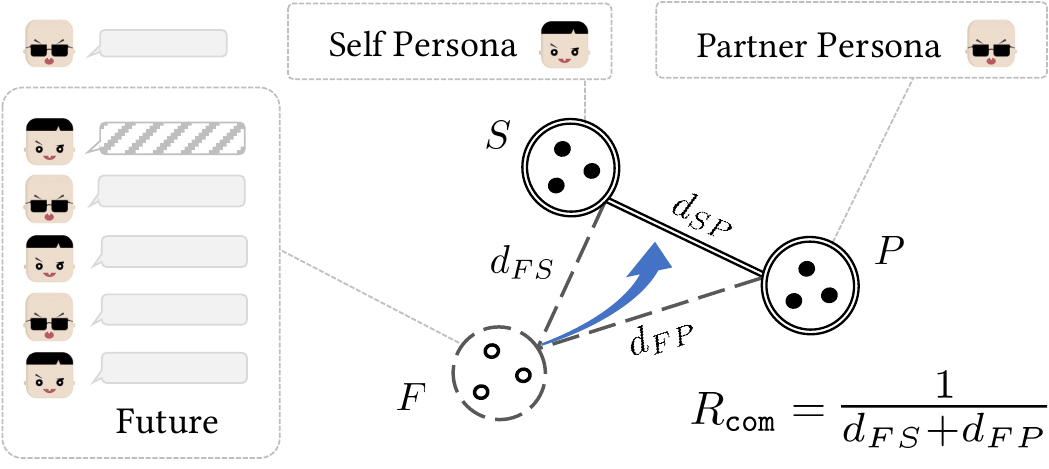}
\vspace{-1.7mm}
\caption{Common Ground Reward. The striped response is optimized as an example.
In this concept set version of triangle $\triangle FSP$, $d_{SP}$ is fixed because of the global concept distance matrix and predefined persona taken from dataset (Eq.~\ref{eq:distance}). However, generated dialogues vary with different model parameters, which results in a "movable" $F$ with two optimizable adjacent sides.
According to the \textit{Triangle Inequality Theorem} that
 $d_{FS} + d_{FP}\geq d_{SP}$, the reward optimize $d_{FS} + d_{FP}$ to the lower bound $d_{SP}$ pushing three concept sets ``collinear''.}
\label{fig:common}
\end{figure}

\begin{table*}[!t]
	\centering
	\caption{Automatic evaluation results on \textsc{Persona-Chat}. The best models are bold and second best ones are underlined within each metric. Baselines are categorized into retrieval based,  generative based, and pre-training \& fine-tuning based methods.}
	\resizebox{.9\textwidth}{!}{
	\begin{tabular}{llcccccc}
		\toprule
		\multirow{2}{*}{\bf Type}&\multirow{2}{*}{\bf{Model}} &
		\multicolumn{3}{c}{\bf Original} &
		\multicolumn{3}{c}{\bf Revised} \cr
		\cmidrule(lr){3-5} \cmidrule(lr){6-8}
		& & Hits@1(\%)$\uparrow$ & F1(\%) $\uparrow$ & Perplexity $\downarrow$ & Hits@1(\%) $\uparrow$ & F1(\%) $\uparrow$ & Perplexity $\downarrow$  \cr
		\cmidrule(lr){1-8}
		\multirow{2}{*}{Retrieval Based}
		
		    &KV Profile Memory~\cite{zhang2018personalizing}& 54.8 & 14.25 & - &38.1&13.65&-  \cr
		    &Dually Interactive Matching~\cite{gu2019dually}& 78.8 & - & -  &\underline{70.7} & - & -  \cr

		\cmidrule(lr){1-8}
		\multirow{3}{*}{Generative Based}
		    &LSTM~\cite{zhang2018personalizing} & - & 16.30 & 50.67  & -& 13.59 & 51.61  \cr
    		&Seq2Seq Attention~\cite{bahdanau2015neural}& 12.5 & 16.82 & 35.07 & 15.52 & 16.82 & 39.54 \cr
    		&Generative Profile Memory~\cite{zhang2018personalizing}& 10.2 & 16.29 & 35.01 & 9.9 & 15.71 & 34.94 \cr

		\cmidrule(lr){1-8}
		\multirow{5}{*}{\vtop{\hbox{\strut Pre-training}\hbox{\strut \& Fine-tuning}\hbox{\strut Based}}}
			&GPT-2 ~\cite{radford2019language} & 18.1 & 18.79 & 17.07  & 17.6& 18.11 & 19.98  \cr
    		&Lost In Conversation~\cite{dinan2019the}& 17.3 & 17.79 & -&16.2 & 16.83 & -  \cr
    		&Transfertransfo~\cite{wolf2019transfertransfo} & \underline{82.1} & 19.09 & 17.51  & - & - & -  \cr
    		&$\mathcal{P}^2$ Bot~\cite{liu2020you}& 81.9 & \underline{19.77} & \bf15.12 &68.6 & \bf 19.08 & \bf18.89  \cr
    		&\modelname{} (Ours)&\bf{85.5} &\bf{20.16} & \underline{16.77} &\bf{74.4} & \underline{18.79} & \underline{19.92}  \cr
		\bottomrule
	\end{tabular}}
	\label{tab:auto}
	\vspace{-1mm}

\end{table*}

\subsubsection{\modelname{} via Policy Gradient}
In order to generate the future dialogue, we adopt the self-play~\cite{lewis2017deal, zhong2021keyword, liu2020you}. That is, we random pair another \modelname{} as the partner to complete a dialogue together. The partner whose parameters are frozen starts the conversation, and the trainable one generates response (we call the trainable one as self in contrast to the partner). This process repeats $N$-turns keeping the conversation flowing (note that the first utterance is directly taken
from the dataset and all $N$ self-utterances are optimized simultaneously). Three elements of reinforcement environment are defined as follows: \textcircled{\raisebox{-0.844pt}{1}} State: the input to the model ($U_n, P^S$); \textcircled{\raisebox{-0.844pt}{2}} Action: the generated response ${u}^S_{n}$; \textcircled{\raisebox{-0.844pt}{3}} Policy: the self \modelname{} $P_{\theta}(u^S_{n}|U_n, P^S)$. Then the policy gradient~\cite{sutton1999policy} is used to optimize the self \modelname{} with the expected return:
\begin{equation}
    J(\theta) =  \mathds E_{{u}^S_n \sim P_{\theta}(u^S_n|U_n, P^S)} [R(u^S_n)]
\end{equation}
With the log derivative trick, we get the gradient of the policy performance, namely expected return:
\begin{equation}
    \begin{aligned}
        \nabla_{\theta} J(\theta) =  \mathds E_{{u}^S_n \sim P_{\theta}(u^S_n|U_n, P^S)} [\nabla_{\theta}\texttt{log}P_{\theta}({u}^S_n|U_n, P^S)R({u}^S_n)]
    \end{aligned}
\end{equation}
which can be further estimated with a batch of samples mean.
In the above equation, the final reward $R$ is the combination of three:
\begin{equation}
    R = \beta_1 R_{\texttt{LM}}+ \beta_2 R_{\texttt{mut}} +\beta_3 R_{\texttt{com}}
\end{equation}
where the $R_{\texttt{LM}}$ is the log likelihood score given by the GPT to keep the text fluency because the other two rewards are somewhat concept-oriented, and 
the Mutual Benefit Reward $R_{\texttt{mut}}$ and Common Ground Reward $R_{\texttt{com}}$ are defined in the next sections.

\subsubsection{Mutual Benefit Reward}
\label{sec:mbr}

This reward uses the combined effects of persona recall score and utterance coherent score to stimulate model curiosity about the partner and the willing to give the partner more
opportunities to express. In addition, the recall score individually can also encourage the model to copy both persona concepts into the responses. We define the reward as follows:
\begin{equation}
    R_{\texttt{mut}} = \gamma S_{\texttt{rec}} + (1-\gamma) S_{\texttt{coh}}
\end{equation}
The recall score of future dialogue is calculated as shown in Figure \ref{fig:recall}.
We first get the intersection concept set of the future dialogue and both party personas by the intersection operation (Eq.~\ref{eq:intersection}):
\begin{equation}
    \begin{aligned}
        \mathbf{c}^{F\cap (S\cup P)} &= \texttt{Inter}\texttt{(} \mathbf{c}^{F} \texttt{,}\ \mathbf{c}^{S\cup P} \texttt{;} r \texttt{)}\\
    \end{aligned}
\end{equation}
Then the recall score can be calculated by the following equation:
\begin{equation}
    S_{\texttt{rec}} = \frac{\lVert \mathbf{c}^{F\cap (S\cup P)} \rVert_1}{\lVert \mathbf{c}^{S\cup P}\rVert_1}
\end{equation}
The coherent score is the mean of the scores following previous utterance and followed by later utterance given by the classifier:
\begin{equation}
        S_{\texttt{coh}} = (C_{u_n^{S}} + C_{u_{n+1}^{P}})/2\\
\end{equation}
\begin{equation}
    \begin{aligned}
        C_{u_n^{S}} &= \texttt{log}P(\texttt{IsNext}|u^{S}_{n}, P^{S}, U_n)\\
        C_{u_{n+1}^{P}} &= \texttt{log}P(\texttt{IsNext}|u^{P}_{n+1}, P^{P}, [U_{n}, u_n^S]) 
    \end{aligned}
\end{equation}


\subsubsection{Common Ground Reward}
\label{sec:cgr}
This reward aims to encourage the model to find common interests, explore new concepts close to both parties, and maintain the topic around the common field. Specifically, the reward is designed from a geometrical perspective by viewing both personas as two constraints and simultaneously minimizing the concept set distances from the future dialogue to each of them. As shown in Figure \ref{fig:common}, the distance from the concept set of future dialogue $F$ to the both persona sets ($S$ and $P$) can be calculated by the set distance operation (Eq.~\ref{eq:distance}).
Then the reward $R_\texttt{com}$ aims to minimize the sum of these two concept set distances:
\begin{equation}
    \begin{aligned}
    R_{\texttt{com}} &= \frac{1}{\texttt{Dist}(\mathbf{c}^{F}, \mathbf{c}^{S}) + \texttt{Dist}(\mathbf{c}^{F},
    \mathbf{c}^{P})}
    \end{aligned}
\end{equation}

\begin{equation}\label{eq:ds}
    \texttt{Dist}(\mathbf{c}^{F}, \mathbf{c}^{S})  = \frac{{(\mathbf c^F)}^{\top}}{\lVert \mathbf c^F\rVert_1} \cdot \mathbf{D} \cdot \frac{\mathbf c^S}{\lVert \mathbf c^S\rVert_1}
\end{equation}

\begin{equation}\label{eq:ds}
    \texttt{Dist}(\mathbf{c}^{F},
    \mathbf{c}^{P}) = \frac{{(\mathbf c^F)}^{\top}}{\lVert \mathbf c^F\rVert_1} \cdot \mathbf{D} \cdot \frac{\mathbf c^P}{\lVert \mathbf c^P\rVert_1}
\end{equation}


\section{Experimental Setup}
\label{sec:experiments}

\subsection{Research Questions}
We list the research questions we want to investigate in this paper:
\begin{itemize}[leftmargin=6mm]
    \item \textbf{RQ1}: Can \modelname{} make a good performance in both automatic and human evaluations by generating responses consistent with personas across both parties? (See \hyperref[sec:eval]{\textsection\ref{sec:eval}})
    \item \textbf{RQ2}: How do different key components contribute to moving the generated responses towards the ground truth? (See \hyperref[sec:eval]{\textsection\ref{sec:abl}})
    \item \textbf{RQ3}: How is \modelname{} guided by concept set during the generation? How do different modules work together to build responses?  (See \hyperref[sec:eval]{\textsection\ref{sec:guide}})
    \item \textbf{RQ4}: Can our rewards really stimulate the model's curiosity and give the partner more chances to express? (See \hyperref[sec:eval]{\textsection\ref{sec:cur}})
    \item \textbf{RQ5}: Can our rewards really improve the model's attention to the partner when generating responses? (See \hyperref[sec:eval]{\textsection\ref{sec:beh}})
    \item \textbf{RQ6}: What is the influence of generalization strength of concept sets (parameter $k$ and $r$) on the performance? (See \hyperref[sec:eval]{\textsection\ref{sec:eff}})
    \item \textbf{RQ7}: What is the dfferences of response pattern between ours and baselines? (See \hyperref[sec:eval]{\textsection\ref{sec:case}})

\end{itemize}
We answer the above questions in the results and analysis section.

\begin{table}[t]
	\caption{Human evaluation Results}
	\centering
	\resizebox{0.9\columnwidth}{!}{
	\begin{tabular}{lllll}
		\toprule
		Models & Fluency & Engagement & Consistency & Avg. \cr
		\cmidrule(lr){1-5}
	    TransferTransfo      & 4.43 & 3.64 & 3.83 & 3.97 \cr
	    $\mathcal P^{2}$ Bot & \bf{4.57} & 3.98 & 4.31 & 4.29  \cr
	    \modelname{}          & 4.52 & \bf{4.35} & \bf{4.37} & \bf{4.41} \cr
		\bottomrule
	\end{tabular}}
	\label{tab:human}
\end{table}

\begin{figure*}[!t]
\centering
\includegraphics[width=1.01\textwidth]{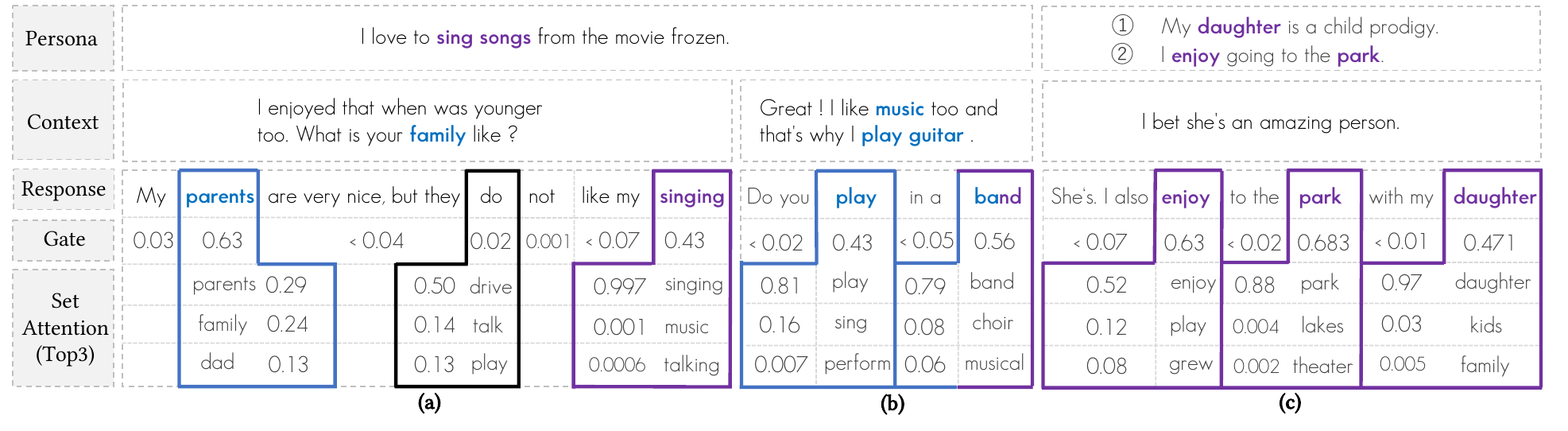}
\vspace{-7.5mm}
\caption{Visualization of concept set guided generation across both personas (a, b) and multiple persona sentences (c). The gate, top-3 concept attention probabilities, and related persona sentences are presented. We show the time steps when \modelname{} copies concepts related to self persona (\textcolor{purple}{purple}), partner persona in context (\textcolor{blue}{blue}), and both parties (e.g. ``\textcolor{blue}{ba}\textcolor{purple}{nd}''). We also show a time when the gate is closed (e.g. ``do'').}
\label{fig:vis}
\end{figure*}

\subsection{Datasets}
We conduct experiments on the \textsc{Persona-Chat}~\cite{zhang2018personalizing}
where each persona is described with at least 5 profile sentences. The dataset contains 8,939/1,000 multi-turn dialogues conditioned on 1,155/100 personas for train/dev. We report all results on the dev set. We also present the automatic results on revised \textsc{Persona-Chat} where the original personas are rephrased, generalized, or specialized because there is a danger that during the dataset construction, humans will unwittingly copy persona information either verbatim or with significant word overlap. This may make the task less challenging and constrain the generalization ability of models~\cite{zhang2018personalizing}.

\subsection{Evaluation Metrics}
\subsubsection{Automatic Evaluation}
Following~\citet{zhang2018personalizing}, three common metrics are used: F1, Hits@1/20 and Perplexity (PPL). F1 is calculated from the word-level precision and recall between the gold response and generated response.
Hits@1 is the accuracy that measures whether the model distinguishes the gold from distracting ones. Here, 19 distracting responses are mixed with one gold response. 
PPL evaluates the generation quality of language models.
\subsubsection{Human Evaluation}
Three crowdsource workers are asked to evaluate Fluency (1-5), Engagingness (1-5) and Consistency (1-5). For engagingness, 1 point means that the response is boring and general. 3 point means that it is a interesting, informative and unexpected but reasonable response. 5 means that it has all the properties of 3 points responses and at the meantime, the dialogue is conducted around the common ground with interactivity. Consistency measures whether the generated response is consistent with its own persona. 1 is not ``consistent'', 3 is ``consistent'' and 5 represents the response is consistent with self-persona while also considers partners' persona. That is, the response as a bridge to mention both personas can achieve a good score. For example, when meeting a music student, a good response for a computer student should be ``Hi, would you like to hear some \textbf{music} created by my \textbf{computer programs}?'' instead of saying a lot of computer knowledge though these contents are consistent with self-persona.

\subsection{Baseline Methods}
We compare our model with three groups of highly correlated and strong baselines: retrieval, generative, and pre-training \& fine-tuning based methods. Specifically, they are introduced as follows:
\begin{itemize}[leftmargin=4mm]
    \item Retrival Based: KV Profile Memory is proposed by \citet{zhang2018personalizing}, a key-value memory neural model by taking as input the dialogue history and calculate attention over the persona sentences along with the history; Dually Interactive Matching Network (DIM) proposed by \citet{gu2019dually} with interactive matching between the context and response as well as between the persona and response simultaneously. 
    \item Generative Based: LSTM language model is trained by prepending persona to the input sequence. Generative Profile Memory Network is based on RNN to encode persona texts into memory representations. Both are proposed by \citet{zhang2018personalizing}. Seq2Seq Attention is built by the traditional attention-based method~\cite{bahdanau2015neural}.
    \item Pre-traing \& Fine-tuning Based: GPT-2~\cite{radford2019language} was fine-tuned on the dataset with concatenated persona and history as prefix. Lost in Conversation team ~\cite{dinan2019the} proposed a conversational model based on the transformer architecture and transfer learning, to pre-train the model on a separate large dataset and fine-tune for the Persona-Chat. \citet{wolf2019transfertransfo} proposed TransferTransfo, a multi-layer transformer ~\cite{vaswani2017attention} based on the Generative Pre-trained Transformer (GPT) ~\cite{radford2018improving}, pre-trained and then fine-tuned with fully supervised learning (SL);
    $\mathcal{P}^2$ Bot was proposed by \citet{liu2020you} who fine-tuned a pre-trained language model by a receiver trained on negative sampling to enhance mutual persona understanding.
\end{itemize}

\begin{table}[t]
	\vspace{-4mm}
	\centering
	\caption{Ablation study on \textsc{Persona-Chat}}
	\vspace{-2.5mm}
	\resizebox{.9\columnwidth}{!}{
	\begin{tabular}{lll}
		\toprule
		Variant & F1(\%) $\uparrow$ & BLEU (\%) $\uparrow$ \cr
		\cmidrule(lr){1-3}
	    \modelname{} Base & 19.25 & 0.94\cr
	    - Concept Copy Mechanism & 19.09 ($-$0.8\%)& 0.89  ($-$5.3\%)\cr
	    + Language Model Reward & 19.28 ($+$0.1\%) & 0.95 ($+$1.1\%)\cr
	    $\hookrightarrow$ + Mutual Benefit Reward & 19.58 ($+$2.0\%) & 1.04  ($+$9.5\%)\cr
	    \ \ \ \ \ \    $\hookrightarrow$ + Common Ground Reward & 20.16 ($+$3.0\%) & 1.10 ($+$5.8\%)\cr
		\bottomrule
	\end{tabular}}
	\label{tab:ablation}
\end{table}

\subsection{Implementation Details}
All concept set operations are based on the matrix computations implemented by by Pytorch~\cite{paszke2019pytorch}. Two parameters of set operations $k$ and $r$ are set to 250 and 0.2 respectively and the length of concept vocabulary $V$ is 2600. Our model is built on GPT-2~\cite{radford2019language} of HuggingFace transformers~\cite{wolf2019huggingface} and run on one single NIVDIA V100 GPU. In addition, the self-play framework is modified from ~\citet{ miller-etal-2017-parlai} and~\citet{liu2020you}. We also thank the open source codes from ~\citet{zhong2021keyword} and ~\citet{ji2020language} for references. For hyper-parameters, all $\alpha_1$, $\alpha_2$, $\beta_1$, $\beta_2$, $\beta_3$ are set as 1 and $\gamma$ as 0.5. Before we add all rewards together, we perform batch normalization to push them into the same 0-1 scale and batch standardization to only view a half rewards as positive ones to encourage and the others as negative ones. The number of turns of future dialogue is 3, the batch size is 10/6 for supervised fine-tuning/reinforcement fine-training, and the beam size is 2.

\section{Results and Analysis}

\subsection{Automatic and Human Evaluations (RQ1)}
\label{sec:eval}

\subsubsection{Automatec Evaluation} 
We present the automatic evaluation results 
in Table \ref{tab:auto}. We can see that
in general, our model achieves either the best or the second best performance across all metrics and outperforms all baselines on original F1 and Hits@1 while presents competitive performance on PPL. The results indicate that by giving more attention to the partner, our model can generate more human-like responses (F1$\uparrow$). In addition, based on the concept set framework, our model can better recognize the gold response among the distractors, even in situations where the persona is revised (8.5\% improvement on Hits@1 against the strongest baseline).

\subsubsection{Human Evaluation}
As automated metrics are notoriously poor for evaluating dialogue~\cite{liu2016how}, we also perform the human evaluation. We collect 384 generated responses for each model from 50 different sampled dialogues. 
We report the averaged scores of them and the results are shown in Table \ref{tab:human}. Three models obtain similar scores on fluency while our model significantly (Wilcoxon signed-rank test~\cite{unknown} is used with $p<0.05$) outperforms all the baselines on engagingness and consistency.


\subsection{Ablation Study (RQ2)}
\label{sec:abl}
We conduct ablation study to verify how different components contribute to the performance as shown in Table \ref{tab:ablation}. We use F1 and BLEU as metrics to measure how close between the generated responses and the ground truths. \modelname{} Base denotes the version of \modelname{} only fine-tuned by the supervised phase, with the gate control to copy concepts from both personas at the appropriate time. Dropping concept copy mechanism means we shut up the guide from the concept set, which results in deteriorated performance indicating the importance of the guidance from the concepts. In addition, Mutual Benefit Reward and Common Ground Reward contribute to the largest improvements in F1 and BLEU respectively, demonstrating that being curious about the partner, leading conversation around mutual personas, and finding the common ground plays an important role in pushing response towards the ground truth of how a conversation really goes.

\begin{figure}[!t]
\vspace{-3.6mm}
\centering
\includegraphics[width=\columnwidth]{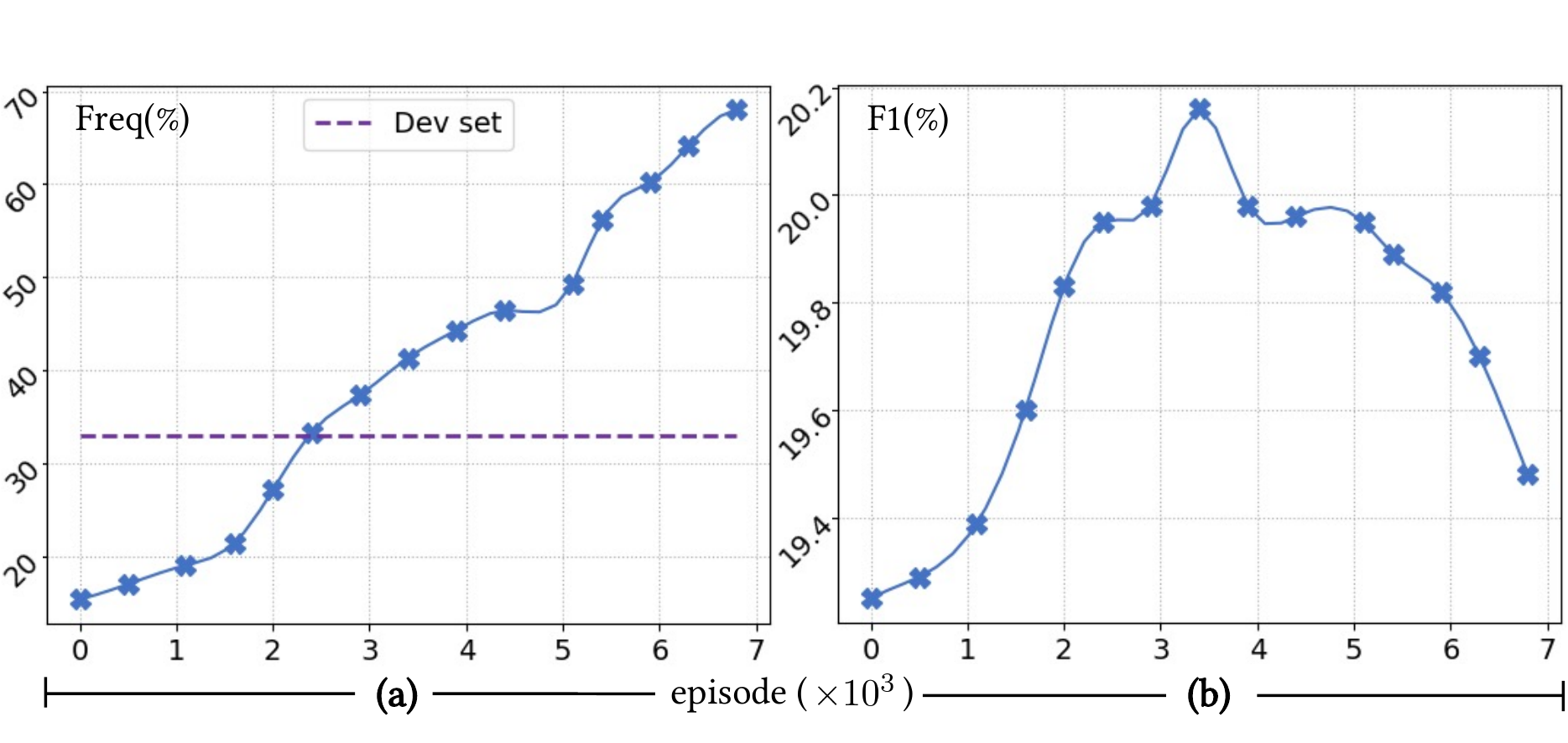}
\vspace{-6mm}
\caption{The response behaviour change through reinforcement learning. 
We measure \modelname{} Base (a) adding questions by the frequency of more than one question mark in the response as well as (b) the performance change. One episode corresponds to one self-play dialogue.}
\label{fig:doublef}
\end{figure}

\subsection{Effectiveness of Concept Set Guided Response Generation (RQ3)}
\label{sec:guide}
We demonstrate how \modelname{} is guided by the concept set during the generation in Figure~\ref{fig:vis} and give the following observations:
\begin{enumerate}[leftmargin=5mm]
    \item The concept copy module (\hyperref[sec:ccm]{\textsection\ref{sec:ccm}}) is able to find out the appropriate time to open the gate.
    \item The concept attention module (\hyperref[sec:ca]{\textsection\ref{sec:ca}}) independently prepares the best concepts with the correct grammar even at the time step when the gate is closed (e.g. ``do'').
    \item \modelname{} is able to generate response across both personas by two ways. The first way is to bridge both-party concepts (e.g., ``parents'' and ``singing'' in Figure~\ref{fig:vis}a). The other way is to copy common ground concepts which is not directly covered by both personas while bridges them (e.g., ``music''/``guitar'' $\rightarrow$ ``band'' $\leftarrow$ ``sing''/``songs'' in Figure~\ref{fig:vis}b).
    \item Not only across both parties, but \modelname{} is also able to copy concepts across multiple persona sentences (Fig~\ref{fig:vis}c).
\end{enumerate}
The above observations show the more human-like and sophisticated ability of \modelname{} to organize multiple information across both parties and multiple persona sentences into one response.

\begin{figure}[!t]
\centering
\includegraphics[width=\columnwidth]{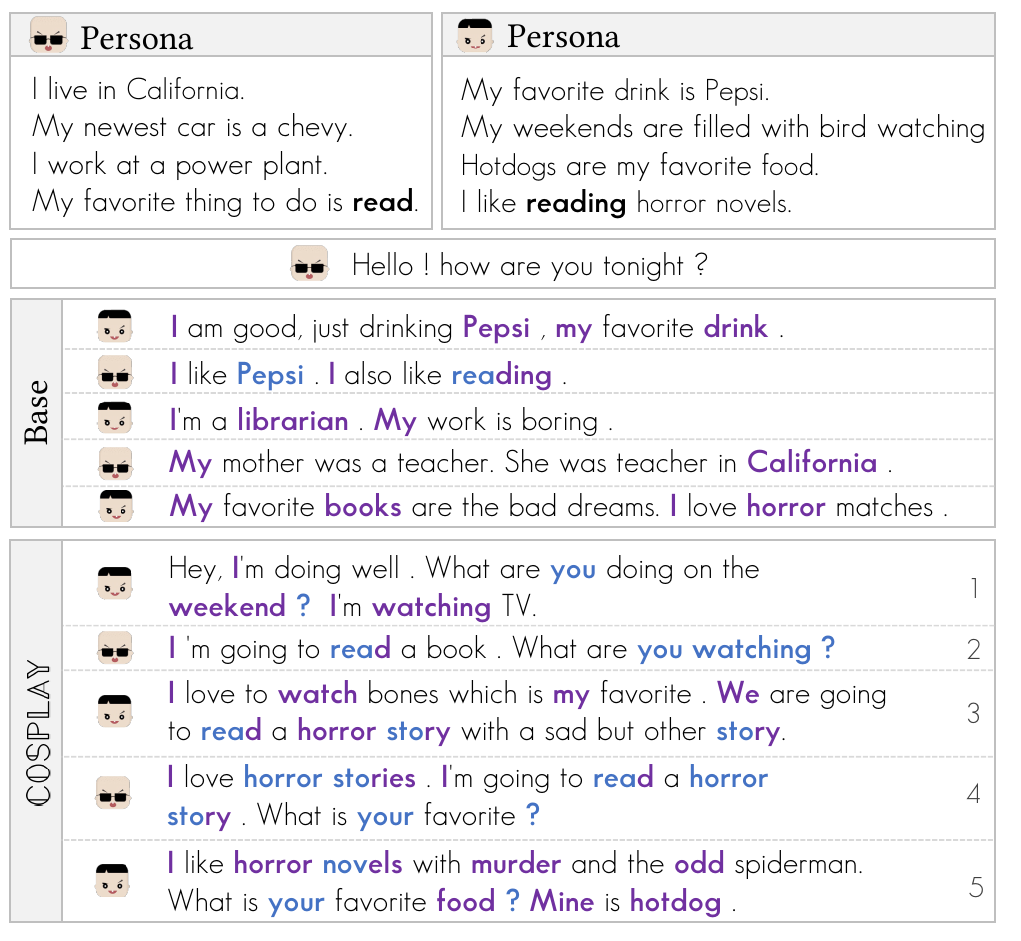} 
\caption{Attention distribution over both parties in two self-play dialogues are made before (Base) and after reinforcement fine-training (\modelname{}) respectively. The settings in these two dialogues are same (same two personas and same first utterance taken from the dataset). The bold words in both personas are the common ground concepts,
words in \textcolor{purple}{purple} represent the attention to itself (first-person pronouns \& self-persona related concepts), words in \textcolor{blue}{blue} denotes the attention on the partner (second-person pronouns \& question mark \& partner persona related concepts), and words in hybrid color (e.g. ``\textcolor{blue}{rea}\textcolor{purple}{d}'', ``\textcolor{blue}{sto}\textcolor{purple}{ry}'', ``\textcolor{blue}{nov}\textcolor{purple}{els}'') denotes the concepts related to the common ground.}
\label{fig:behaviour}
\end{figure}

\subsection{Effectiveness of Encouraging Curiosity for the Partner (RQ4)}

\label{sec:cur}

\label{sec:behaviour}


To better observe whether our rewards can stimulate the model’s curiosity, we deliberately train the model with a specified episodes without early stopping (Figure \ref{fig:doublef}). It's clearly that the power of Mutual Benefit Reward is great. Without interruption, the \modelname{} tends to fuse question for every response through the RL in a generative way (Figure \ref{fig:doublef}a). However, asking question too frequently can make the performance decrease (Figure \ref{fig:doublef}b). After all in daily life, though frequently, people may not ask questions for every time. In addition, there is a delicate balance of question-asking and performance (Figure \ref{fig:doublef}b). Hence we use F1 to early stop the reinforcement fine-training when the performance declines.


\vspace{+2mm}
\subsection{Effectiveness of Encouraging Balance of Attention between Both Parties (RQ5)}
\label{sec:beh}
Two self-play dialogues in Figure \ref{fig:behaviour} are compared to evaluate our rewards on the effect of improving the model's attention to its partner when generating responses. Some key observations are:
\begin{enumerate}[leftmargin=5mm]
    \item The frequency of asking question increases, indicating that the model has learned to balance expressing self-persona and keeping curiosity toward the partner. The way of asking questions can be further categorized: 
    \begin{enumerate}[leftmargin=5mm]
        \item copy partner's persona concept in the question (e.g., ``What are you \textbf{watching}?'' in line 2): a common way to know more about the partner by giving them more chances to express persona information; 
        \item copy self-persona concept in the question (e.g., ``What are you doing on the \textbf{weekend}?'' in line 1 and ``What is your favorite food? Mine is \textbf{hotdog}.'' in line 5): a common way to find common ground by proposing our interests first.
    \end{enumerate}
    \item More concepts related to "read" appear in the dialogue (e.g. ``read'', ``story'', and ``novel''), demonstrating that the model has learned to find common interests, explore new concepts close to both parties, and maintain the topic around common field.
    \item The length of responses increases and the proportion of two colors becomes more balanced, which shows that our model has learned to balance speaking and listening.
\end{enumerate}

\subsection{Effects of Generalization Strength of Concept Set (RQ6)}
\label{sec:eff}
\begin{figure}[!t]
\vspace{-3.7mm}
\centering
\includegraphics[width=\columnwidth]{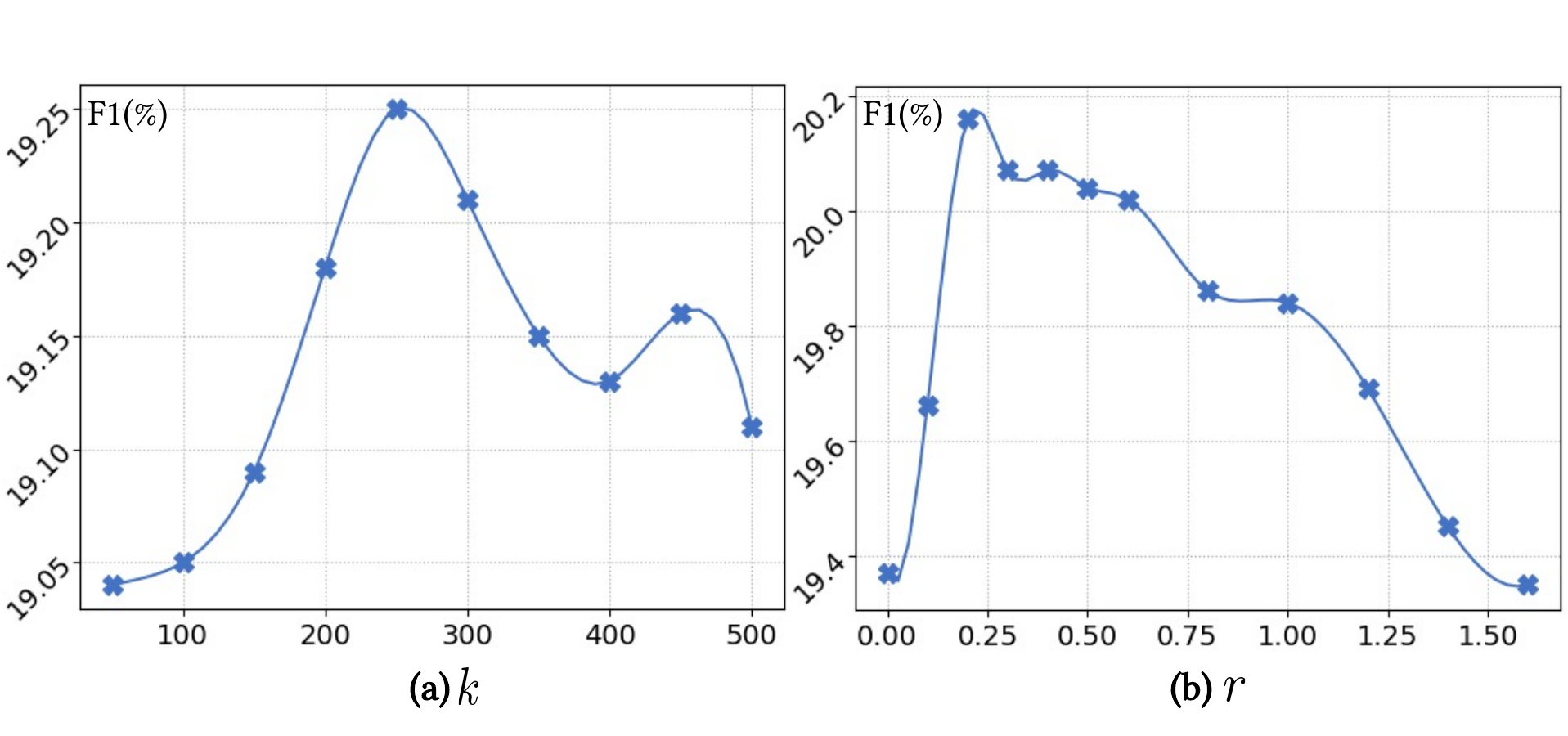}
\vspace{-8mm}
\caption{Performance with different values of operation parameters: $k$ of set expansion and $r$ of set intersection.}
\label{fig:hyper}
\end{figure}

Two parameters controls the generalization strength as shown in Figure~\ref{fig:hyper}: 1) $k$ decides the number of concepts used to guide generation. Concept set expansion operation will fill concepts from near to far until the number of the set reaches $k$. 2) $r$ decides how close two concepts can they considered as ``same''. If $r$ approaches to zero, then the concept set intersection operation will degenerate into the traditional one.
It is clearly that both without and over generalizing concepts decrease the performance.

\subsection{Case Study (RQ7)}
\label{sec:case}
Figure \ref{fig:case} shows three generated responses compared with our two baselines: TransferTransfo (SOTA 1) and $\mathcal P^2$ Bot (SOTA 2). Based on the analyses demonstrated in the Introduction and the section \hyperref[sec:guide]{\textsection\ref{sec:guide}}, we conclude that by improving attention to the partner, our model is able to generate less egocentric and more human-like responses.

\subsection{Limitations}
One major limitation of the Concept Set framework is that all concepts considered are single content words, which sometimes makes the \modelname{} separately process the phrase. For example, as shown in Figure~\ref{fig:behaviour}, the model says ``I'm watching TV'' (line 1) where the word ``watching'' is partly copied from the phrase ``bird watching'' to construct the response. A potential solution is to extend the framework into its phrase version. In addition, to further take advantage of the matrix calculation feature of the Concept Set framework, making the concept distance matrix trainable leaves another research direction.

\section{Conclusion}

We propose \modelname{} with \textbf{Concept Set} framework to address the problem of egocentrism in the personalized dialogue generation. Concept Set framework enables us to model personas and dialogues all in the form of concept sets, as well as using a suite of knowledge-enhanced concept set operations to calculate their relationships. Based on Concept Set, the concepts of both party personas, the concept relationships between them, and their relationships to the future dialogues are all used to train the model in the supervised fine-tuning and reinforcement fine-training phases, in order to give more attention to its partner and give partner more opportunities to express persona during generating responses. Experiments on \textsc{Persona-Chat} demonstrate that \modelname{} is capable of generating less egocentric, more human-like, and higher quality responses.


\section{Acknowledgments}
We thank the anonymous reviewers whose suggestions helped clarify this work. This research
is supported in part by the National Natural Science Foundation of China (Grant No. 62106105).
We would like to thank Dr. Piji Li and Prof. Chuangbai Xiao for insightful suggestions and careful mentoring.
We also want to thank Dr. Yan Wang, Dr. Wei Bi, and other members of the NLP group at Tencent AI Lab for helpful discussions.
\bibliographystyle{ACM-Reference-Format}
\balance
\bibliography{main}


\begin{thebibliography}{47}


\ifx \showCODEN    \undefined \def \showCODEN     #1{\unskip}     \fi
\ifx \showDOI      \undefined \def \showDOI       #1{#1}\fi
\ifx \showISBNx    \undefined \def \showISBNx     #1{\unskip}     \fi
\ifx \showISBNxiii \undefined \def \showISBNxiii  #1{\unskip}     \fi
\ifx \showISSN     \undefined \def \showISSN      #1{\unskip}     \fi
\ifx \showLCCN     \undefined \def \showLCCN      #1{\unskip}     \fi
\ifx \shownote     \undefined \def \shownote      #1{#1}          \fi
\ifx \showarticletitle \undefined \def \showarticletitle #1{#1}   \fi
\ifx \showURL      \undefined \def \showURL       {\relax}        \fi
\providecommand\bibfield[2]{#2}
\providecommand\bibinfo[2]{#2}
\providecommand\natexlab[1]{#1}
\providecommand\showeprint[2][]{arXiv:#2}

\bibitem[\protect\citeauthoryear{{Bahdanau}, {Cho}, and {Bengio}}{{Bahdanau}
  et~al\mbox{.}}{2015}]%
        {bahdanau2015neural}
\bibfield{author}{\bibinfo{person}{Dzmitry {Bahdanau}},
  \bibinfo{person}{Kyunghyun {Cho}}, {and} \bibinfo{person}{Yoshua {Bengio}}.}
  \bibinfo{year}{2015}\natexlab{}.
\newblock \showarticletitle{Neural Machine Translation by Jointly Learning to
  Align and Translate}. In \bibinfo{booktitle}{\emph{ICLR 2015 : International
  Conference on Learning Representations 2015}}.
\newblock


\bibitem[\protect\citeauthoryear{Cao, Bi, Fang, Shi, and Tao}{Cao
  et~al\mbox{.}}{2022}]%
        {bi2022amodel}
\bibfield{author}{\bibinfo{person}{Yu Cao}, \bibinfo{person}{Wei Bi},
  \bibinfo{person}{Meng Fang}, \bibinfo{person}{Shuming Shi}, {and}
  \bibinfo{person}{Dacheng Tao}.} \bibinfo{year}{2022}\natexlab{}.
\newblock \bibinfo{title}{A Model-Agnostic Data Manipulation Method for
  Persona-based Dialogue Generation}.
\newblock
\newblock
\urldef\tempurl%
\url{https://doi.org/10.48550/ARXIV.2204.09867}
\showDOI{\tempurl}


\bibitem[\protect\citeauthoryear{{Dinan}, {Logacheva}, {Malykh}, {Miller},
  {Shuster}, {Urbanek}, {Kiela}, {Szlam}, {Serban}, {Lowe}, {Prabhumoye},
  {Black}, {Rudnicky}, {Williams}, {Pineau}, {Burtsev}, and {Weston}}{{Dinan}
  et~al\mbox{.}}{2019}]%
        {dinan2019the}
\bibfield{author}{\bibinfo{person}{Emily {Dinan}}, \bibinfo{person}{Varvara
  {Logacheva}}, \bibinfo{person}{Valentin {Malykh}},
  \bibinfo{person}{Alexander~H. {Miller}}, \bibinfo{person}{Kurt {Shuster}},
  \bibinfo{person}{Jack {Urbanek}}, \bibinfo{person}{Douwe {Kiela}},
  \bibinfo{person}{Arthur {Szlam}}, \bibinfo{person}{Iulian {Serban}},
  \bibinfo{person}{Ryan {Lowe}}, \bibinfo{person}{Shrimai {Prabhumoye}},
  \bibinfo{person}{Alan~W. {Black}}, \bibinfo{person}{Alexander~I. {Rudnicky}},
  \bibinfo{person}{Jason {Williams}}, \bibinfo{person}{Joelle {Pineau}},
  \bibinfo{person}{Mikhail~S. {Burtsev}}, {and} \bibinfo{person}{Jason
  {Weston}}.} \bibinfo{year}{2019}\natexlab{}.
\newblock \showarticletitle{The Second Conversational Intelligence Challenge
  (ConvAI2)}.
\newblock \bibinfo{journal}{\emph{arXiv preprint arXiv:1902.00098}}
  (\bibinfo{year}{2019}), \bibinfo{pages}{187--208}.
\newblock


\bibitem[\protect\citeauthoryear{Durango and Refugio}{Durango and
  Refugio}{2018}]%
        {unknown}
\bibfield{author}{\bibinfo{person}{Ana Durango} {and} \bibinfo{person}{Craig
  Refugio}.} \bibinfo{year}{2018}\natexlab{}.
\newblock \bibinfo{title}{An Empirical Study on Wilcoxon Signed Rank Test}.
\newblock
\newblock
\urldef\tempurl%
\url{https://doi.org/10.13140/RG.2.2.13996.51840}
\showDOI{\tempurl}


\bibitem[\protect\citeauthoryear{Golovanov, Kurbanov, Nikolenko, Truskovskyi,
  Tselousov, and Wolf}{Golovanov et~al\mbox{.}}{2019}]%
        {golovanov2019large}
\bibfield{author}{\bibinfo{person}{Sergey Golovanov}, \bibinfo{person}{Rauf
  Kurbanov}, \bibinfo{person}{Sergey Nikolenko}, \bibinfo{person}{Kyryl
  Truskovskyi}, \bibinfo{person}{Alexander Tselousov}, {and}
  \bibinfo{person}{Thomas Wolf}.} \bibinfo{year}{2019}\natexlab{}.
\newblock \showarticletitle{Large-scale transfer learning for natural language
  generation}. In \bibinfo{booktitle}{\emph{Proceedings of the 57th Annual
  Meeting of the Association for Computational Linguistics}}.
  \bibinfo{pages}{6053--6058}.
\newblock


\bibitem[\protect\citeauthoryear{{Gu}, {Lu}, {Li}, and {Li}}{{Gu}
  et~al\mbox{.}}{2016}]%
        {gu2016incorporating}
\bibfield{author}{\bibinfo{person}{Jiatao {Gu}}, \bibinfo{person}{Zhengdong
  {Lu}}, \bibinfo{person}{Hang {Li}}, {and} \bibinfo{person}{Victor~O.K.
  {Li}}.} \bibinfo{year}{2016}\natexlab{}.
\newblock \showarticletitle{Incorporating Copying Mechanism in
  Sequence-to-Sequence Learning}. In \bibinfo{booktitle}{\emph{Proceedings of
  the 54th Annual Meeting of the Association for Computational Linguistics
  (Volume 1: Long Papers)}}, Vol.~\bibinfo{volume}{1}.
  \bibinfo{pages}{1631--1640}.
\newblock


\bibitem[\protect\citeauthoryear{{Gu}, {Ling}, {Zhu}, and {Liu}}{{Gu}
  et~al\mbox{.}}{2019}]%
        {gu2019dually}
\bibfield{author}{\bibinfo{person}{Jia-Chen {Gu}}, \bibinfo{person}{Zhen-Hua
  {Ling}}, \bibinfo{person}{Xiaodan {Zhu}}, {and} \bibinfo{person}{Quan
  {Liu}}.} \bibinfo{year}{2019}\natexlab{}.
\newblock \showarticletitle{Dually Interactive Matching Network for
  Personalized Response Selection in Retrieval-Based Chatbots.}. In
  \bibinfo{booktitle}{\emph{Proceedings of the 2019 Conference on Empirical
  Methods in Natural Language Processing and the 9th International Joint
  Conference on Natural Language Processing (EMNLP-IJCNLP)}}.
  \bibinfo{pages}{1845--1854}.
\newblock


\bibitem[\protect\citeauthoryear{Gu, Liu, Ling, Liu, Chen, and Zhu}{Gu
  et~al\mbox{.}}{2021}]%
        {gu2021partner}
\bibfield{author}{\bibinfo{person}{Jia-Chen Gu}, \bibinfo{person}{Hui Liu},
  \bibinfo{person}{Zhen-Hua Ling}, \bibinfo{person}{Quan Liu},
  \bibinfo{person}{Zhigang Chen}, {and} \bibinfo{person}{Xiaodan Zhu}.}
  \bibinfo{year}{2021}\natexlab{}.
\newblock \showarticletitle{Partner Matters! An Empirical Study on Fusing
  Personas for Personalized Response Selection in Retrieval-Based Chatbots}.
\newblock \bibinfo{journal}{\emph{arXiv preprint arXiv:2105.09050}}
  (\bibinfo{year}{2021}).
\newblock


\bibitem[\protect\citeauthoryear{Guan, Wang, and Huang}{Guan
  et~al\mbox{.}}{2019}]%
        {guan2019story}
\bibfield{author}{\bibinfo{person}{Jian Guan}, \bibinfo{person}{Yansen Wang},
  {and} \bibinfo{person}{Minlie Huang}.} \bibinfo{year}{2019}\natexlab{}.
\newblock \showarticletitle{Story ending generation with incremental encoding
  and commonsense knowledge}. In \bibinfo{booktitle}{\emph{Proceedings of the
  AAAI Conference on Artificial Intelligence}}, Vol.~\bibinfo{volume}{33}.
  \bibinfo{pages}{6473--6480}.
\newblock


\bibitem[\protect\citeauthoryear{{Ji}, {Ke}, {Huang}, {Wei}, {Zhu}, and
  {Huang}}{{Ji} et~al\mbox{.}}{2020}]%
        {ji2020language}
\bibfield{author}{\bibinfo{person}{Haozhe {Ji}}, \bibinfo{person}{Pei {Ke}},
  \bibinfo{person}{Shaohan {Huang}}, \bibinfo{person}{Furu {Wei}},
  \bibinfo{person}{Xiaoyan {Zhu}}, {and} \bibinfo{person}{Minlie {Huang}}.}
  \bibinfo{year}{2020}\natexlab{}.
\newblock \showarticletitle{Language Generation with Multi-Hop Reasoning on
  Commonsense Knowledge Graph.}. In \bibinfo{booktitle}{\emph{Proceedings of
  the 2020 Conference on Empirical Methods in Natural Language Processing
  (EMNLP)}}. \bibinfo{pages}{725--736}.
\newblock


\bibitem[\protect\citeauthoryear{{Kottur}, {Wang}, and {Carvalho}}{{Kottur}
  et~al\mbox{.}}{2017}]%
        {kottur2017exploring}
\bibfield{author}{\bibinfo{person}{Satwik {Kottur}}, \bibinfo{person}{Xiaoyu
  {Wang}}, {and} \bibinfo{person}{Vitor~R. {Carvalho}}.}
  \bibinfo{year}{2017}\natexlab{}.
\newblock \showarticletitle{Exploring personalized neural conversational
  models}. In \bibinfo{booktitle}{\emph{IJCAI'17 Proceedings of the 26th
  International Joint Conference on Artificial Intelligence}}.
  \bibinfo{pages}{3728--3734}.
\newblock


\bibitem[\protect\citeauthoryear{{Lewis}, {Yarats}, {Dauphin}, {Parikh}, and
  {Batra}}{{Lewis} et~al\mbox{.}}{2017}]%
        {lewis2017deal}
\bibfield{author}{\bibinfo{person}{Mike {Lewis}}, \bibinfo{person}{Denis
  {Yarats}}, \bibinfo{person}{Yann~N. {Dauphin}}, \bibinfo{person}{Devi
  {Parikh}}, {and} \bibinfo{person}{Dhruv {Batra}}.}
  \bibinfo{year}{2017}\natexlab{}.
\newblock \showarticletitle{Deal or No Deal? End-to-End Learning of Negotiation
  Dialogues}. In \bibinfo{booktitle}{\emph{Proceedings of the 2017 Conference
  on Empirical Methods in Natural Language Processing}}.
  \bibinfo{pages}{2443--2453}.
\newblock


\bibitem[\protect\citeauthoryear{Li, Galley, Brockett, Spithourakis, Gao, and
  Dolan}{Li et~al\mbox{.}}{2016}]%
        {li2016persona}
\bibfield{author}{\bibinfo{person}{Jiwei Li}, \bibinfo{person}{Michel Galley},
  \bibinfo{person}{Chris Brockett}, \bibinfo{person}{Georgios~P Spithourakis},
  \bibinfo{person}{Jianfeng Gao}, {and} \bibinfo{person}{Bill Dolan}.}
  \bibinfo{year}{2016}\natexlab{}.
\newblock \showarticletitle{A persona-based neural conversation model}.
\newblock \bibinfo{journal}{\emph{arXiv preprint arXiv:1603.06155}}
  (\bibinfo{year}{2016}).
\newblock


\bibitem[\protect\citeauthoryear{Li, Li, Ren, Ren, and Chen}{Li
  et~al\mbox{.}}{2020}]%
        {li2020knowledge}
\bibfield{author}{\bibinfo{person}{Qintong Li}, \bibinfo{person}{Piji Li},
  \bibinfo{person}{Zhaochun Ren}, \bibinfo{person}{Pengjie Ren}, {and}
  \bibinfo{person}{Zhumin Chen}.} \bibinfo{year}{2020}\natexlab{}.
\newblock \bibinfo{title}{Knowledge Bridging for Empathetic Dialogue
  Generation}.
\newblock
\newblock
\urldef\tempurl%
\url{https://doi.org/10.48550/ARXIV.2009.09708}
\showDOI{\tempurl}


\bibitem[\protect\citeauthoryear{{Liu}, {Lowe}, {Serban}, {Noseworthy},
  {Charlin}, and {Pineau}}{{Liu} et~al\mbox{.}}{2016}]%
        {liu2016how}
\bibfield{author}{\bibinfo{person}{Chia-Wei {Liu}}, \bibinfo{person}{Ryan
  {Lowe}}, \bibinfo{person}{Iulian~Vlad {Serban}}, \bibinfo{person}{Michael
  {Noseworthy}}, \bibinfo{person}{Laurent {Charlin}}, {and}
  \bibinfo{person}{Joelle {Pineau}}.} \bibinfo{year}{2016}\natexlab{}.
\newblock \showarticletitle{How NOT To Evaluate Your Dialogue System: An
  Empirical Study of Unsupervised Evaluation Metrics for Dialogue Response
  Generation}. In \bibinfo{booktitle}{\emph{Proceedings of the 2016 Conference
  on Empirical Methods in Natural Language Processing}}.
  \bibinfo{pages}{2122--2132}.
\newblock


\bibitem[\protect\citeauthoryear{{Liu}, {Chen}, {Chen}, {Lou}, {Chen}, {Zhou},
  and {Zhang}}{{Liu} et~al\mbox{.}}{2020}]%
        {liu2020you}
\bibfield{author}{\bibinfo{person}{Qian {Liu}}, \bibinfo{person}{Yihong
  {Chen}}, \bibinfo{person}{Bei {Chen}}, \bibinfo{person}{Jian-Guang {Lou}},
  \bibinfo{person}{Zixuan {Chen}}, \bibinfo{person}{Bin {Zhou}}, {and}
  \bibinfo{person}{Dongmei {Zhang}}.} \bibinfo{year}{2020}\natexlab{}.
\newblock \showarticletitle{You Impress Me: Dialogue Generation via Mutual
  Persona Perception}. In \bibinfo{booktitle}{\emph{Proceedings of the 58th
  Annual Meeting of the Association for Computational Linguistics}}.
  \bibinfo{pages}{1417--1427}.
\newblock


\bibitem[\protect\citeauthoryear{{Mazaré}, {Humeau}, {Raison}, and
  {Bordes}}{{Mazaré} et~al\mbox{.}}{2018}]%
        {mazar2018training}
\bibfield{author}{\bibinfo{person}{Pierre-Emmanuel {Mazaré}},
  \bibinfo{person}{Samuel {Humeau}}, \bibinfo{person}{Martin {Raison}}, {and}
  \bibinfo{person}{Antoine {Bordes}}.} \bibinfo{year}{2018}\natexlab{}.
\newblock \showarticletitle{Training Millions of Personalized Dialogue Agents}.
  In \bibinfo{booktitle}{\emph{Proceedings of the 2018 Conference on Empirical
  Methods in Natural Language Processing}}. \bibinfo{pages}{2775--2779}.
\newblock


\bibitem[\protect\citeauthoryear{Miller, Feng, Batra, Bordes, Fisch, Lu,
  Parikh, and Weston}{Miller et~al\mbox{.}}{2017}]%
        {miller-etal-2017-parlai}
\bibfield{author}{\bibinfo{person}{Alexander Miller}, \bibinfo{person}{Will
  Feng}, \bibinfo{person}{Dhruv Batra}, \bibinfo{person}{Antoine Bordes},
  \bibinfo{person}{Adam Fisch}, \bibinfo{person}{Jiasen Lu},
  \bibinfo{person}{Devi Parikh}, {and} \bibinfo{person}{Jason Weston}.}
  \bibinfo{year}{2017}\natexlab{}.
\newblock \showarticletitle{{P}arl{AI}: A Dialog Research Software Platform}.
  In \bibinfo{booktitle}{\emph{Proceedings of the 2017 Conference on Empirical
  Methods in Natural Language Processing: System Demonstrations}}.
  \bibinfo{publisher}{Association for Computational Linguistics},
  \bibinfo{address}{Copenhagen, Denmark}, \bibinfo{pages}{79--84}.
\newblock
\urldef\tempurl%
\url{https://doi.org/10.18653/v1/D17-2014}
\showDOI{\tempurl}


\bibitem[\protect\citeauthoryear{Paszke, Gross, Massa, Lerer, Bradbury, Chanan,
  Killeen, Lin, Gimelshein, Antiga, et~al\mbox{.}}{Paszke
  et~al\mbox{.}}{2019}]%
        {paszke2019pytorch}
\bibfield{author}{\bibinfo{person}{Adam Paszke}, \bibinfo{person}{Sam Gross},
  \bibinfo{person}{Francisco Massa}, \bibinfo{person}{Adam Lerer},
  \bibinfo{person}{James Bradbury}, \bibinfo{person}{Gregory Chanan},
  \bibinfo{person}{Trevor Killeen}, \bibinfo{person}{Zeming Lin},
  \bibinfo{person}{Natalia Gimelshein}, \bibinfo{person}{Luca Antiga},
  {et~al\mbox{.}}} \bibinfo{year}{2019}\natexlab{}.
\newblock \showarticletitle{Pytorch: An imperative style, high-performance deep
  learning library}.
\newblock \bibinfo{journal}{\emph{Advances in neural information processing
  systems}}  \bibinfo{volume}{32} (\bibinfo{year}{2019}),
  \bibinfo{pages}{8026--8037}.
\newblock


\bibitem[\protect\citeauthoryear{Qian, Huang, Zhao, Xu, and Zhu}{Qian
  et~al\mbox{.}}{2018}]%
        {persona-huang}
\bibfield{author}{\bibinfo{person}{Qiao Qian}, \bibinfo{person}{Minlie Huang},
  \bibinfo{person}{Haizhou Zhao}, \bibinfo{person}{Jingfang Xu}, {and}
  \bibinfo{person}{Xiaoyan Zhu}.} \bibinfo{year}{2018}\natexlab{}.
\newblock \showarticletitle{Assigning Personality/Profile to a Chatting Machine
  for Coherent Conversation Generation}. In
  \bibinfo{booktitle}{\emph{Proceedings of the 27th International Joint
  Conference on Artificial Intelligence}} (Stockholm, Sweden)
  \emph{(\bibinfo{series}{IJCAI'18})}. \bibinfo{publisher}{AAAI Press},
  \bibinfo{pages}{4279–4285}.
\newblock
\showISBNx{9780999241127}


\bibitem[\protect\citeauthoryear{{Qian}, {Huang}, {Zhao}, {Xu}, and
  {Zhu}}{{Qian} et~al\mbox{.}}{2018}]%
        {qian2018assigning}
\bibfield{author}{\bibinfo{person}{Qiao {Qian}}, \bibinfo{person}{Minlie
  {Huang}}, \bibinfo{person}{Haizhou {Zhao}}, \bibinfo{person}{Jingfang {Xu}},
  {and} \bibinfo{person}{Xiaoyan {Zhu}}.} \bibinfo{year}{2018}\natexlab{}.
\newblock \showarticletitle{Assigning Personality/Profile to a Chatting Machine
  for Coherent Conversation Generation}. In
  \bibinfo{booktitle}{\emph{Proceedings of the Twenty-Seventh International
  Joint Conference on Artificial Intelligence}}. \bibinfo{pages}{4279--4285}.
\newblock


\bibitem[\protect\citeauthoryear{Radford, Narasimhan, Salimans, and
  Sutskever}{Radford et~al\mbox{.}}{2018}]%
        {radford2018improving}
\bibfield{author}{\bibinfo{person}{Alec Radford}, \bibinfo{person}{Karthik
  Narasimhan}, \bibinfo{person}{Tim Salimans}, {and} \bibinfo{person}{Ilya
  Sutskever}.} \bibinfo{year}{2018}\natexlab{}.
\newblock \showarticletitle{Improving language understanding by generative
  pre-training}.
\newblock  (\bibinfo{year}{2018}).
\newblock


\bibitem[\protect\citeauthoryear{Radford, Wu, Child, Luan, Amodei, and
  Sutskever}{Radford et~al\mbox{.}}{2019}]%
        {radford2019language}
\bibfield{author}{\bibinfo{person}{Alec Radford}, \bibinfo{person}{Jeff Wu},
  \bibinfo{person}{Rewon Child}, \bibinfo{person}{David Luan},
  \bibinfo{person}{Dario Amodei}, {and} \bibinfo{person}{Ilya Sutskever}.}
  \bibinfo{year}{2019}\natexlab{}.
\newblock \showarticletitle{Language Models are Unsupervised Multitask
  Learners}.
\newblock  (\bibinfo{year}{2019}).
\newblock


\bibitem[\protect\citeauthoryear{{See}, {Liu}, and {Manning}}{{See}
  et~al\mbox{.}}{2017}]%
        {see2017get}
\bibfield{author}{\bibinfo{person}{Abigail {See}}, \bibinfo{person}{Peter~J.
  {Liu}}, {and} \bibinfo{person}{Christopher~D. {Manning}}.}
  \bibinfo{year}{2017}\natexlab{}.
\newblock \showarticletitle{Get To The Point: Summarization with
  Pointer-Generator Networks}. In \bibinfo{booktitle}{\emph{Proceedings of the
  55th Annual Meeting of the Association for Computational Linguistics (Volume
  1: Long Papers)}}, Vol.~\bibinfo{volume}{1}. \bibinfo{pages}{1073--1083}.
\newblock


\bibitem[\protect\citeauthoryear{Shang, Lu, and Li}{Shang
  et~al\mbox{.}}{2015}]%
        {shang2015neural}
\bibfield{author}{\bibinfo{person}{Lifeng Shang}, \bibinfo{person}{Zhengdong
  Lu}, {and} \bibinfo{person}{Hang Li}.} \bibinfo{year}{2015}\natexlab{}.
\newblock \showarticletitle{Neural Responding Machine for Short-Text
  Conversation}. In \bibinfo{booktitle}{\emph{Proceedings of the 53rd Annual
  Meeting of the Association for Computational Linguistics and the 7th
  International Joint Conference on Natural Language Processing (Volume 1: Long
  Papers)}}. \bibinfo{pages}{1577--1586}.
\newblock


\bibitem[\protect\citeauthoryear{Shum, He, and Li}{Shum et~al\mbox{.}}{2018}]%
        {shum2018eliza}
\bibfield{author}{\bibinfo{person}{Heung-Yeung Shum}, \bibinfo{person}{Xiaodong
  He}, {and} \bibinfo{person}{Di Li}.} \bibinfo{year}{2018}\natexlab{}.
\newblock \showarticletitle{From Eliza to XiaoIce: challenges and opportunities
  with social chatbots}.
\newblock \bibinfo{journal}{\emph{arXiv preprint arXiv:1801.01957}}
  (\bibinfo{year}{2018}).
\newblock


\bibitem[\protect\citeauthoryear{Song, Wang, Zhang, Zhang, and Liu}{Song
  et~al\mbox{.}}{2021}]%
        {song-etal-2021-bob}
\bibfield{author}{\bibinfo{person}{Haoyu Song}, \bibinfo{person}{Yan Wang},
  \bibinfo{person}{Kaiyan Zhang}, \bibinfo{person}{Wei-Nan Zhang}, {and}
  \bibinfo{person}{Ting Liu}.} \bibinfo{year}{2021}\natexlab{}.
\newblock \showarticletitle{{B}o{B}: {BERT} Over {BERT} for Training
  Persona-based Dialogue Models from Limited Personalized Data}. In
  \bibinfo{booktitle}{\emph{Proceedings of the 59th Annual Meeting of the
  Association for Computational Linguistics and the 11th International Joint
  Conference on Natural Language Processing (Volume 1: Long Papers)}}.
  \bibinfo{publisher}{Association for Computational Linguistics},
  \bibinfo{address}{Online}, \bibinfo{pages}{167--177}.
\newblock
\urldef\tempurl%
\url{https://doi.org/10.18653/v1/2021.acl-long.14}
\showDOI{\tempurl}


\bibitem[\protect\citeauthoryear{Song, Wang, Zhang, Liu, and Liu}{Song
  et~al\mbox{.}}{2020a}]%
        {song2020generate}
\bibfield{author}{\bibinfo{person}{Haoyu Song}, \bibinfo{person}{Yan Wang},
  \bibinfo{person}{Wei-Nan Zhang}, \bibinfo{person}{Xiaojiang Liu}, {and}
  \bibinfo{person}{Ting Liu}.} \bibinfo{year}{2020}\natexlab{a}.
\newblock \showarticletitle{Generate, delete and rewrite: A three-stage
  framework for improving persona consistency of dialogue generation}.
\newblock \bibinfo{journal}{\emph{arXiv preprint arXiv:2004.07672}}
  (\bibinfo{year}{2020}).
\newblock


\bibitem[\protect\citeauthoryear{Song, Zhang, Hu, and Liu}{Song
  et~al\mbox{.}}{2020b}]%
        {Song_RCDG_2020}
\bibfield{author}{\bibinfo{person}{Haoyu Song}, \bibinfo{person}{Wei-Nan
  Zhang}, \bibinfo{person}{Jingwen Hu}, {and} \bibinfo{person}{Ting Liu}.}
  \bibinfo{year}{2020}\natexlab{b}.
\newblock \showarticletitle{Generating Persona Consistent Dialogues by
  Exploiting Natural Language Inference}.
\newblock \bibinfo{journal}{\emph{Proceedings of the AAAI Conference on
  Artificial Intelligence}} \bibinfo{volume}{34}, \bibinfo{number}{05}
  (\bibinfo{date}{Apr.} \bibinfo{year}{2020}), \bibinfo{pages}{8878--8885}.
\newblock
\urldef\tempurl%
\url{https://doi.org/10.1609/aaai.v34i05.6417}
\showDOI{\tempurl}


\bibitem[\protect\citeauthoryear{Speer, Chin, and Havasi}{Speer
  et~al\mbox{.}}{2017}]%
        {speer2017conceptnet}
\bibfield{author}{\bibinfo{person}{Robyn Speer}, \bibinfo{person}{Joshua Chin},
  {and} \bibinfo{person}{Catherine Havasi}.} \bibinfo{year}{2017}\natexlab{}.
\newblock \showarticletitle{Conceptnet 5.5: An open multilingual graph of
  general knowledge}. In \bibinfo{booktitle}{\emph{Thirty-first AAAI conference
  on artificial intelligence}}.
\newblock


\bibitem[\protect\citeauthoryear{{Sutton}, {McAllester}, {Singh}, and
  {Mansour}}{{Sutton} et~al\mbox{.}}{1999}]%
        {sutton1999policy}
\bibfield{author}{\bibinfo{person}{Richard~S {Sutton}},
  \bibinfo{person}{David~A. {McAllester}}, \bibinfo{person}{Satinder~P.
  {Singh}}, {and} \bibinfo{person}{Yishay {Mansour}}.}
  \bibinfo{year}{1999}\natexlab{}.
\newblock \showarticletitle{Policy Gradient Methods for Reinforcement Learning
  with Function Approximation}. In \bibinfo{booktitle}{\emph{Advances in Neural
  Information Processing Systems 12}}, Vol.~\bibinfo{volume}{12}.
  \bibinfo{pages}{1057--1063}.
\newblock


\bibitem[\protect\citeauthoryear{Tian, Bi, Zhang, Lee, Song, and Zhang}{Tian
  et~al\mbox{.}}{2021}]%
        {tian2021learning}
\bibfield{author}{\bibinfo{person}{Zhiliang Tian}, \bibinfo{person}{Wei Bi},
  \bibinfo{person}{Zihan Zhang}, \bibinfo{person}{Dongkyu Lee},
  \bibinfo{person}{Yiping Song}, {and} \bibinfo{person}{Nevin~L Zhang}.}
  \bibinfo{year}{2021}\natexlab{}.
\newblock \showarticletitle{Learning from My Friends: Few-Shot Personalized
  Conversation Systems via Social Networks}.
\newblock \bibinfo{journal}{\emph{arXiv preprint arXiv:2105.10323}}
  (\bibinfo{year}{2021}).
\newblock


\bibitem[\protect\citeauthoryear{Vaswani, Shazeer, Parmar, Uszkoreit, Jones,
  Gomez, Kaiser, and Polosukhin}{Vaswani et~al\mbox{.}}{2017}]%
        {vaswani2017attention}
\bibfield{author}{\bibinfo{person}{Ashish Vaswani}, \bibinfo{person}{Noam
  Shazeer}, \bibinfo{person}{Niki Parmar}, \bibinfo{person}{Jakob Uszkoreit},
  \bibinfo{person}{Llion Jones}, \bibinfo{person}{Aidan~N Gomez},
  \bibinfo{person}{{\L}ukasz Kaiser}, {and} \bibinfo{person}{Illia
  Polosukhin}.} \bibinfo{year}{2017}\natexlab{}.
\newblock \showarticletitle{Attention is all you need}. In
  \bibinfo{booktitle}{\emph{Advances in neural information processing
  systems}}. \bibinfo{pages}{5998--6008}.
\newblock


\bibitem[\protect\citeauthoryear{Wang, Huang, Xu, Shen, and Nie}{Wang
  et~al\mbox{.}}{2018}]%
        {wang2018chat}
\bibfield{author}{\bibinfo{person}{Wenjie Wang}, \bibinfo{person}{Minlie
  Huang}, \bibinfo{person}{Xin-Shun Xu}, \bibinfo{person}{Fumin Shen}, {and}
  \bibinfo{person}{Liqiang Nie}.} \bibinfo{year}{2018}\natexlab{}.
\newblock \showarticletitle{Chat more: Deepening and widening the chatting
  topic via a deep model}. In \bibinfo{booktitle}{\emph{The 41st international
  acm sigir conference on research \& development in information retrieval}}.
  \bibinfo{pages}{255--264}.
\newblock


\bibitem[\protect\citeauthoryear{{Williams}}{{Williams}}{1992}]%
        {williams1992simple}
\bibfield{author}{\bibinfo{person}{Ronald~J. {Williams}}.}
  \bibinfo{year}{1992}\natexlab{}.
\newblock \showarticletitle{Simple Statistical Gradient-Following Algorithms
  for Connectionist Reinforcement Learning}.
\newblock \bibinfo{journal}{\emph{Machine Learning}} \bibinfo{volume}{8},
  \bibinfo{number}{3} (\bibinfo{year}{1992}), \bibinfo{pages}{229--256}.
\newblock


\bibitem[\protect\citeauthoryear{Wolf, Debut, Sanh, Chaumond, Delangue, Moi,
  Cistac, Rault, Louf, Funtowicz, et~al\mbox{.}}{Wolf et~al\mbox{.}}{2019}]%
        {wolf2019huggingface}
\bibfield{author}{\bibinfo{person}{Thomas Wolf}, \bibinfo{person}{Lysandre
  Debut}, \bibinfo{person}{Victor Sanh}, \bibinfo{person}{Julien Chaumond},
  \bibinfo{person}{Clement Delangue}, \bibinfo{person}{Anthony Moi},
  \bibinfo{person}{Pierric Cistac}, \bibinfo{person}{Tim Rault},
  \bibinfo{person}{R{\'e}mi Louf}, \bibinfo{person}{Morgan Funtowicz},
  {et~al\mbox{.}}} \bibinfo{year}{2019}\natexlab{}.
\newblock \showarticletitle{Huggingface's transformers: State-of-the-art
  natural language processing}.
\newblock \bibinfo{journal}{\emph{arXiv preprint arXiv:1910.03771}}
  (\bibinfo{year}{2019}).
\newblock


\bibitem[\protect\citeauthoryear{{Wolf}, {Sanh}, {Chaumond}, and
  {Delangue}}{{Wolf} et~al\mbox{.}}{2019}]%
        {wolf2019transfertransfo}
\bibfield{author}{\bibinfo{person}{Thomas {Wolf}}, \bibinfo{person}{Victor
  {Sanh}}, \bibinfo{person}{Julien {Chaumond}}, {and} \bibinfo{person}{Clement
  {Delangue}}.} \bibinfo{year}{2019}\natexlab{}.
\newblock \showarticletitle{TransferTransfo: A Transfer Learning Approach for
  Neural Network Based Conversational Agents}.
\newblock \bibinfo{journal}{\emph{arXiv preprint arXiv:1901.08149}}
  (\bibinfo{year}{2019}).
\newblock


\bibitem[\protect\citeauthoryear{Xu, Zhao, Li, Hu, and Xiao}{Xu
  et~al\mbox{.}}{2021}]%
        {xu2021change}
\bibfield{author}{\bibinfo{person}{Chen Xu}, \bibinfo{person}{Jianyu Zhao},
  \bibinfo{person}{Rang Li}, \bibinfo{person}{Changjian Hu}, {and}
  \bibinfo{person}{Chuangbai Xiao}.} \bibinfo{year}{2021}\natexlab{}.
\newblock \showarticletitle{Change or not: A simple approach for plug and play
  language models on sentiment control}. In
  \bibinfo{booktitle}{\emph{Proceedings of the AAAI Conference on Artificial
  Intelligence}}, Vol.~\bibinfo{volume}{35}. \bibinfo{pages}{15935--15936}.
\newblock


\bibitem[\protect\citeauthoryear{Xu, Li, Yang, Ren, Ren, Chen, and Ma}{Xu
  et~al\mbox{.}}{2020}]%
        {xu2020neural}
\bibfield{author}{\bibinfo{person}{Minghong Xu}, \bibinfo{person}{Piji Li},
  \bibinfo{person}{Haoran Yang}, \bibinfo{person}{Pengjie Ren},
  \bibinfo{person}{Zhaochun Ren}, \bibinfo{person}{Zhumin Chen}, {and}
  \bibinfo{person}{Jun Ma}.} \bibinfo{year}{2020}\natexlab{}.
\newblock \showarticletitle{A Neural Topical Expansion Framework for
  Unstructured Persona-Oriented Dialogue Generation}.
\newblock In \bibinfo{booktitle}{\emph{ECAI 2020}}. \bibinfo{publisher}{IOS
  Press}, \bibinfo{pages}{2244--2251}.
\newblock


\bibitem[\protect\citeauthoryear{Yavuz, Rastogi, Chao, and Hakkani-Tur}{Yavuz
  et~al\mbox{.}}{2019}]%
        {yavuz2019deepcopy}
\bibfield{author}{\bibinfo{person}{Semih Yavuz}, \bibinfo{person}{Abhinav
  Rastogi}, \bibinfo{person}{Guan-Lin Chao}, {and} \bibinfo{person}{Dilek
  Hakkani-Tur}.} \bibinfo{year}{2019}\natexlab{}.
\newblock \showarticletitle{Deepcopy: Grounded response generation with
  hierarchical pointer networks}.
\newblock \bibinfo{journal}{\emph{arXiv preprint arXiv:1908.10731}}
  (\bibinfo{year}{2019}).
\newblock


\bibitem[\protect\citeauthoryear{{Zhang}, {Dinan}, {Urbanek}, {Szlam}, {Kiela},
  and {Weston}}{{Zhang} et~al\mbox{.}}{2018}]%
        {zhang2018personalizing}
\bibfield{author}{\bibinfo{person}{Saizheng {Zhang}}, \bibinfo{person}{Emily
  {Dinan}}, \bibinfo{person}{Jack {Urbanek}}, \bibinfo{person}{Arthur {Szlam}},
  \bibinfo{person}{Douwe {Kiela}}, {and} \bibinfo{person}{Jason {Weston}}.}
  \bibinfo{year}{2018}\natexlab{}.
\newblock \showarticletitle{Personalizing Dialogue Agents: I have a dog, do you
  have pets too?}. In \bibinfo{booktitle}{\emph{Proceedings of the 56th Annual
  Meeting of the Association for Computational Linguistics (Volume 1: Long
  Papers)}}, Vol.~\bibinfo{volume}{1}. \bibinfo{pages}{2204--2213}.
\newblock


\bibitem[\protect\citeauthoryear{Zhang, Sun, Galley, Chen, Brockett, Gao, Gao,
  Liu, and Dolan}{Zhang et~al\mbox{.}}{2020}]%
        {zhang2019dialogpt}
\bibfield{author}{\bibinfo{person}{Yizhe Zhang}, \bibinfo{person}{Siqi Sun},
  \bibinfo{person}{Michel Galley}, \bibinfo{person}{Yen-Chun Chen},
  \bibinfo{person}{Chris Brockett}, \bibinfo{person}{Xiang Gao},
  \bibinfo{person}{Jianfeng Gao}, \bibinfo{person}{Jingjing Liu}, {and}
  \bibinfo{person}{Bill Dolan}.} \bibinfo{year}{2020}\natexlab{}.
\newblock \showarticletitle{DialoGPT: Large-Scale Generative Pre-training for
  Conversational Response Generation}. In \bibinfo{booktitle}{\emph{ACL, system
  demonstration}}.
\newblock


\bibitem[\protect\citeauthoryear{Zhao, Zhao, and Eskenazi}{Zhao
  et~al\mbox{.}}{2017}]%
        {zhao2017learning}
\bibfield{author}{\bibinfo{person}{Tiancheng Zhao}, \bibinfo{person}{Ran Zhao},
  {and} \bibinfo{person}{Maxine Eskenazi}.} \bibinfo{year}{2017}\natexlab{}.
\newblock \showarticletitle{Learning Discourse-level Diversity for Neural
  Dialog Models using Conditional Variational Autoencoders}. In
  \bibinfo{booktitle}{\emph{Proceedings of the 55th Annual Meeting of the
  Association for Computational Linguistics (Volume 1: Long Papers)}}.
  \bibinfo{pages}{654--664}.
\newblock


\bibitem[\protect\citeauthoryear{Zheng, Zhang, Huang, and Mao}{Zheng
  et~al\mbox{.}}{2020}]%
        {zheng2020pre}
\bibfield{author}{\bibinfo{person}{Yinhe Zheng}, \bibinfo{person}{Rongsheng
  Zhang}, \bibinfo{person}{Minlie Huang}, {and} \bibinfo{person}{Xiaoxi Mao}.}
  \bibinfo{year}{2020}\natexlab{}.
\newblock \showarticletitle{A pre-training based personalized dialogue
  generation model with persona-sparse data}. In
  \bibinfo{booktitle}{\emph{Proceedings of the AAAI Conference on Artificial
  Intelligence}}, Vol.~\bibinfo{volume}{34}. \bibinfo{pages}{9693--9700}.
\newblock


\bibitem[\protect\citeauthoryear{Zhong, Liu, Wang, and Miao}{Zhong
  et~al\mbox{.}}{2021}]%
        {zhong2021keyword}
\bibfield{author}{\bibinfo{person}{Peixiang Zhong}, \bibinfo{person}{Yong Liu},
  \bibinfo{person}{Hao Wang}, {and} \bibinfo{person}{Chunyan Miao}.}
  \bibinfo{year}{2021}\natexlab{}.
\newblock \showarticletitle{Keyword-Guided Neural Conversational Model}. In
  \bibinfo{booktitle}{\emph{Proceedings of the AAAI Conference on Artificial
  Intelligence}}, Vol.~\bibinfo{volume}{35}. \bibinfo{pages}{14568--14576}.
\newblock


\bibitem[\protect\citeauthoryear{Zhou, Young, Huang, Zhao, Xu, and Zhu}{Zhou
  et~al\mbox{.}}{2018}]%
        {zhou2018commonsense}
\bibfield{author}{\bibinfo{person}{Hao Zhou}, \bibinfo{person}{Tom Young},
  \bibinfo{person}{Minlie Huang}, \bibinfo{person}{Haizhou Zhao},
  \bibinfo{person}{Jingfang Xu}, {and} \bibinfo{person}{Xiaoyan Zhu}.}
  \bibinfo{year}{2018}\natexlab{}.
\newblock \showarticletitle{Commonsense knowledge aware conversation generation
  with graph attention.}. In \bibinfo{booktitle}{\emph{IJCAI}}.
  \bibinfo{pages}{4623--4629}.
\newblock


\bibitem[\protect\citeauthoryear{Zhou, Gopalakrishnan, Hedayatnia, Kim, Pujara,
  Ren, Liu, and Hakkani-Tur}{Zhou et~al\mbox{.}}{2021}]%
        {zhou2021commonsense}
\bibfield{author}{\bibinfo{person}{Pei Zhou}, \bibinfo{person}{Karthik
  Gopalakrishnan}, \bibinfo{person}{Behnam Hedayatnia},
  \bibinfo{person}{Seokhwan Kim}, \bibinfo{person}{Jay Pujara},
  \bibinfo{person}{Xiang Ren}, \bibinfo{person}{Yang Liu}, {and}
  \bibinfo{person}{Dilek Hakkani-Tur}.} \bibinfo{year}{2021}\natexlab{}.
\newblock \showarticletitle{Commonsense-Focused Dialogues for Response
  Generation: An Empirical Study}.
\newblock \bibinfo{journal}{\emph{arXiv preprint arXiv:2109.06427}}
  (\bibinfo{year}{2021}).
\newblock


\end{thebibliography}

\end{document}